\documentclass[conference]{IEEEtran}
\IEEEoverridecommandlockouts
\usepackage{cite}
\usepackage{amsmath,amssymb,amsfonts}
\usepackage{algorithm}
\usepackage{algorithmic}
\usepackage{graphicx}
\usepackage{textcomp}
\usepackage{xcolor}
\usepackage{enumitem}
\usepackage{url}
\usepackage{tabularx}
\usepackage{booktabs}
\usepackage{multirow}
\usepackage{orcidlink}

\usepackage[OT1]{fontenc} 
\AtBeginEnvironment{algorithm}{\footnotesize}
\newcommand{\Dev}{\mathrm{Dev}}
\newcommand{\Cos}{\mathrm{Cos}}
\newcommand{\Dis}{\mathrm{Dis}}
\newcommand{\Avg}{\mathrm{Avg}}
\newcommand{\mcF}{\mathcal{F}}
\newcommand{\zeros}{\mathrm{zeros}}
\DeclareMathOperator{\argmin}{\mathrm{argmin}}
 
\def\BibTeX{{\rm B\kern-.05em{\sc i\kern-.025em b}\kern-.08em
    T\kern-.1667em\lower.7ex\hbox{E}\kern-.125emX}}

\usepackage{comment}
\begin{document}

\title{Dynamic Free-Rider Detection in Cross-Silo Federated Learning via Simulated Attack Patterns\\
%{
    %\footnotesize \textsuperscript{*}Note: Sub-titles are not captured for https://ieeexplore.ieee.org and should not be used}
%\thanks{Identify applicable funding agency here. If none, delete this.}
}

\author{\IEEEauthorblockN{Motoki Nakamura\orcidlink{0009-0004-3894-5023}}
\IEEEauthorblockA{\textit{Fujitsu Limited} \\
Kawasaki, Japan \\
nakamura-motoki@fujitsu.com}

}
\iffalse

\author{
\IEEEauthorblockN{Anonymous Author}
}
\fi

\maketitle

\begin{abstract}
\iffalse
\fi
Federated learning (FL) enables multiple clients to collaboratively train a global model by aggregating local updates without sharing private data. 
In this work, we focus on cross-silo FL, where each client typically represents an independent organization.
However, cross-silo FL can face the challenge of free-riders, clients who submit fake model parameters without performing actual training to obtain the global model without contributing.
Chen et al. proposed a free-rider detection method based on the weight evolving frequency (WEF) of model parameters. 
This detection approach is practical because it requires neither a proxy dataset nor pre-training.
Nevertheless, it struggles to detect ``dynamic'' free-riders who behave honestly in early rounds and later switch to free-riding, particularly under global-model-mimicking attacks such as the delta weight attack and our newly proposed \emph{adaptive WEF-camouflage attack}.
In this paper, we propose a novel detection method \emph{S2-WEF} that simulates the WEF patterns of potential global-model-mimicking attacks on the server side using previously broadcast global models, and identifies clients whose submitted WEF patterns resemble the simulated ones.
To handle a variety of free-rider attack strategies, S2-WEF further combines this simulation-based similarity score with a deviation score computed from mutual comparisons among submitted WEFs, and separates benign and free-rider clients by two-dimensional clustering and per-score classification.
This method enables dynamic detection of clients that transition into free-riders during training without proxy datasets or pre-training.
We conduct extensive experiments across four datasets and five attack types, demonstrating that S2-WEF provides robust dynamic free-rider detection across diverse settings.
\end{abstract}


\section{Introduction}\label{sec:intro}
Federated learning (FL) has emerged as a collaborative framework in which
multiple clients jointly train a global model without directly sharing their
local training data \cite{FedAvg}.
In typical FL, a central server coordinates training by aggregating model
updates submitted by clients and broadcasting the updated global model.
In this work, we focus on \emph{cross-silo FL}, where each client typically
represents an independent organization rather than an individual device.
A fundamental challenge in this setting is that the server cannot directly
verify whether each client has honestly performed local training.
Consequently, a client may avoid local training, submit fake updates,
and still benefit from the global model; such a client is referred to as a
\emph{free-rider}.

Free-riding is particularly relevant to cross-silo FL because participating
organizations may have different costs, benefits, and incentives.
Prior work has pointed out that cross-silo FL may involve business competitors
and long-term collaboration as local data evolve, creating incentives for
free-riding \cite{CrossSiloFL_FR}.
A recent interview study of cross-silo FL stakeholders also identified
fairness of client contributions and incentives as practical concerns; in
particular, clients with smaller or lower-quality datasets may obtain relatively
large benefits from the collaboratively trained model \cite{KuoCrossSilo}.
Moreover, cross-silo FL requires continued computation, system maintenance,
and coordination among organizations.
Although current deployments often rely on established organizational
relationships and legal contracts, such arrangements do not technically
verify whether the agreed local training is actually performed
\cite{KuoCrossSilo}.
These characteristics make reliable verification of client contributions an
important issue in practical cross-silo FL.

Importantly, a client's behavior may change during a long-running FL
collaboration.
Changes in computational budgets, operational priorities, local data,
privacy concerns, or competitive relationships can cause a client that
initially performs local training honestly to later reduce or stop its
contribution while continuing to receive the global model.
%We refer to such behavior as \emph{dynamic free-rider}, in contrast to
%where a client is a \emph{static free-rider}, free-riding throughout all global
%communication rounds.
We refer to such a client as a \emph{dynamic free-rider}, in contrast to
a \emph{static free-rider} that free-rides throughout all global
communication rounds.
Furthermore, as also noted in \cite{ZeTFRi}, dynamic free-riders are more
difficult to detect; a strategic client may behave honestly in early
global communication rounds and later switch to free-riding to evade
detection.
Reliable detection of free-riders, particularly those whose behavior changes
over time, is therefore important for sustaining cross-silo collaboration.

Motivated by this need, numerous free-rider detection methods have been
proposed
\cite{freerider-survey,STD-DAGMM,FRAD,ZeTFRi,CFFL,WEF-defense}.
However, several methods require pre-training on benign client behavior
\cite{STD-DAGMM,FRAD,ZeTFRi} or a server-side proxy dataset
\cite{CFFL,ZeTFRi}, which introduces additional assumptions for practical
cross-silo deployment.
Moreover, existing approaches often assume static free-riders or use
information accumulated across global communication rounds, making behavioral
changes difficult to capture.
WEF-defense \cite{WEF-defense} is practical in that it requires neither
pre-training nor a proxy dataset, but it accumulates Weight Evolving
Frequency (WEF) information across rounds.
Consequently, WEF patterns generated during earlier honest participation can
mask subsequent free-riding behavior.
As shown in Sec.~\ref{sec:SFRonWEF}, this problem is particularly severe for
global-model-mimicking attacks, such as the delta weight attack (DWA)
\cite{STD-DAGMM} and our adaptive WEF-camouflage attack (AWCA).

To overcome these limitations, we propose
\emph{S2-WEF (Submitted and Simulated-Weight Evolving Frequency)}, a
round-wise free-rider detection method for cross-silo FL.
The key idea is to exploit the fact that global-model-mimicking attacks are
constructed from previously broadcast global models.
At each global communication round, the server simulates the WEF pattern
expected from such attacks using past global models and compares it with the
WEF-matrix submitted by each client.
S2-WEF combines this simulation-based similarity score with a deviation score
based on comparisons among submitted WEF-matrices.
It then integrates two-dimensional clustering with score-wise threshold
classification to detect diverse free-rider attacks while suppressing false
positives.
S2-WEF performs detection independently at each global communication round
and requires neither a proxy dataset nor detector pre-training.

Our contributions are as follows:
\begin{enumerate}[label=(\roman*)]
    \item We empirically show that WEF-defense can fail to detect dynamic
    free-riders, particularly under global-model-mimicking attacks.

    \item We introduce AWCA and propose S2-WEF, which detects dynamic
    free-riders on a round-wise basis by combining server-side simulation of
    attack WEF patterns with client-to-client deviation, without requiring a
    proxy dataset or detector pre-training.

    \item We evaluate S2-WEF on four datasets under five free-rider attacks.
    S2-WEF achieves a mean F1-score of at least $0.90$ in 39 of 40 main
    experimental settings and remains robust to different attack scenarios, free-rider ratios, 
    and numbers of clients.
\end{enumerate}
To facilitate reproducible research, the implementation of S2-WEF is publicly available at \url{https://github.com/FujitsuResearch/S2-WEF}.

\section{Preliminaries and Related Studies}\label{sec:prelim}
\subsection{Horizontal Federated Learning}
We focus on horizontal FL, where clients share the same model architecture and feature space but hold different training samples \cite{FLsurvey}.
The objective of horizontal FL is to minimize the average of the client loss $\{f_{i}(w)\}$, where $w$ denotes the parameters of the global model and $f_{i}(w)$ denotes the loss function for the $i$-th client. 
Therefore, the total loss function is defined as $F(w) = \frac{1}{N} \sum_{i=1}^{N} f_i(w)$, and the objective of horizontal FL is defined to be $\argmin_{w} F(w)$. 
%Following FedAvg \cite{FedAvg}, at round $T$ the server updates the global model by averaging uploaded local models:
%\begin{equation}
%w_g^{T+1}=\frac{1}{N}\sum_{i=1}^{N} w_i^{T}.
%\end{equation}
Following FedAvg \cite{FedAvg}, the server updates $w_g^{T+1}=\frac{1}{N}\sum_{i=1}^{N} w_i^{T}$.

Here, we do not use the weighted average based on the number of samples in each dataset, since it is difficult to verify that the reported sample counts are correct in a realistic scenario. 
To avoid cases where learning fails due to false reports, following several previous studies, this paper adopts a method that simply takes the average \cite{STD-DAGMM,WEF-defense,flame}.
%In our experiments under the non-IID setting, the expected value of the number of samples used for training is constant, but the actual number of samples may slightly vary per client due to the sampling algorithm. 

\subsection{Threat Model and Free-Rider Attacks}\label{subsec:threatmodel}
\paragraph{Threat model.}
Free-riders aim to obtain a high-quality global model without contributing to FL.
As they wish to obtain this model as benign clients, we assume that they try to avoid being detected as free riders by the server.
We assume that only one attack type occurs at a time, while multiple free-riders may collude by coordinating their submitted updates.
We also assume that free-riders know the FL protocol (model architecture, loss, learning rate, and aggregation rule), but cannot access or manipulate benign clients' local data.
We consider two types of free-riders: \emph{static} free-riders who never train and free-ride throughout training \cite{WEF-defense}, and \emph{dynamic} free-riders who possess their own data and may behave honestly in early rounds before switching to free-riding (possibly intermittently), making detection more challenging \cite{ZeTFRi}.
In \cite{FRtype}, static free-riders and dynamic free-riders are called anonymous free-riders and selfish free-riders, respectively. 
Overall, our threat model extends \cite{WEF-defense} by explicitly permitting dynamic free-riders.

\paragraph{Honest-majority assumption.}
%We assume the number of free-riders does not exceed half of all clients.
We assume that free-riders constitute strictly fewer than half of all clients.
If free-riders were the majority, the global model would be unlikely to converge or improve, contradicting the free-riders' objective of obtaining a well-trained model.

\paragraph{Defender's capability.}
The server is honest, does not know the number of free-riders in each round, and cannot access clients' local data.
However, it observes all submitted updates and the global model, and can request clients to upload auxiliary information (e.g., WEF-matrices) for detection.

\paragraph{Attack instantiations.}
We consider four existing attacks from prior work \cite{STD-DAGMM,SPA}.
\begin{enumerate}[label=(\roman*)]
    \item \textbf{Random weight attack (RWA)} randomly samples model   updates from a uniform distribution. 
    The free-rider must specify the range of the uniform distribution as $[-R,R]$.
%(ii) \textbf{delta weight attack (DWA)} mimics global progress via $\Delta w_i^{f}=w_g^{T}-w_g^{T-1}$;
    \item \textbf{Delta weight attack (DWA)} generates fake model updates by calculating the difference between two previously received global models. 
    Let ${w_{g}}^{T}$ be the global model received in global communication round $T$ and ${w_{g}}^{T-1}$ be the one received in round $T-1$. 
    Then the free-rider updates the model $w_{i}^{f}$ by adding the difference $\Delta w_{i}^{f}$, which is defined by $\Delta{w_{i}}^{f} = {w_{g}}^{T}-{w_{g}}^{T-1}$.
    \item \textbf{Advanced delta weight attack (ADWA)} adds Gaussian noise to the model update generated by DWA. 
    If multiple free-riders use DWA without modification, their model updates will be identical, making detection by the central server easier. 
    %To avoid this, the free-rider calculates the difference of global models in the same way as in the delta weight attack, and adds appropriate noise as follows:
    %$\Delta{w_{i}}^{f} = {w_{g}}^{T}-{w_{g}}^{T-1}+N(0,\sigma)$.
    In order to avoid such a detection, the attacker of ADWA generates the difference of global models in the same way as in DWA, and adds appropriate noise. 
    The free-rider update is then given by $\Delta{w_{i}}^{f} = {w_{g}}^{T}-{w_{g}}^{T-1}+N(0,\sigma)$.
    \item \textbf{Stochastic perturbations attack (SPA)} uses stochastic perturbations to add Gaussian noise to the received global model and returns a fake model update (see \cite{SPA} for details).
\end{enumerate}
%In addition, we introduce \textbf{adaptive WEF-camouflage attack (AWCA)} in Sec.~\ref{sec:SFRonWEF}.
In addition to these existing attacks, we will introduce \textbf{adaptive WEF-camouflage attack (AWCA)} in Sec.~\ref{sec:SFRonWEF}. 
%as a free-rider attack that is difficult to detect using the existing method \cite{WEF-defense}.
\subsection{Free-Rider Detection Methods}
Existing free-rider detection methods in FL can be broadly categorized into anomaly detection on model updates and contribution evaluation \cite{freerider-survey}.

\paragraph{Anomaly detection on model updates.}
A representative line is DAGMM-based detection, initiated by STD-DAGMM \cite{STD-DAGMM}.
Since DAGMM relies on learning a benign representation (e.g., via an autoencoder), DAGMM-based methods require pre-training on benign behavior, which can be impractical in real deployments.
Beyond this shared requirement, several variants further assume additional server-side resources.
For example, FRAD leverages contribution-related side information (e.g., data quality, computational resources, and recommendation relations) \cite{FRAD}, and ZeTFRi requires a proxy dataset for data-quality assessment \cite{ZeTFRi}.

Another practical anomaly detection approach is WEF-defense \cite{WEF-defense}, which detects free-riders by comparing client-submitted WEF-matrices and does not require proxy datasets or pre-training.
However, it targets static free-riders and can fail to detect dynamic free-riders who behave benignly and later switch to free-riding (see Sec.~\ref{sec:SFRonWEF}).

\paragraph{Contribution evaluation.}
CFFL evaluates contributions using a proxy dataset and distributes models of varying quality accordingly \cite{CFFL}.
RFFL uses similarity-based signals for contribution evaluation \cite{RFFL}, but has been reported to be ineffective against stronger attacks such as DWA \cite{WEF-defense}.
%Moreover, contribution evaluation often depends on historical behaviors across rounds, which makes dynamic detection of switching free-riders challenging.
Moreover, contribution evaluation often depends on historical behavior across rounds, making the dynamic detection of free-riders challenging.

In summary, few existing methods simultaneously avoid pre-training and proxy datasets while effectively handling dynamic free-riders.
Our method S2-WEF targets this gap; see Tab.~\ref{tab:summary} for a concise comparison.

\begin{table*}[t]
\caption{Comparative summary of free-rider detection methods.}
\label{tab:summary}
\centering
\scriptsize
\setlength{\tabcolsep}{3.0pt}
\renewcommand{\arraystretch}{1.10}
\begin{tabularx}{\textwidth}{@{}l*{7}{>{\centering\arraybackslash}X}@{}}
\toprule
Algorithm
& STD-DAGMM~\cite{STD-DAGMM}
& FRAD~\cite{FRAD}
& ZeTFRi~\cite{ZeTFRi}
& CFFL~\cite{CFFL}
& RFFL~\cite{RFFL}
& WEF-Defense~\cite{WEF-defense}
& \textbf{S2-WEF (ours)} \\
\midrule
No pre-training
& $\times$ & $\times$ & $\times$ & $\checkmark$ & $\checkmark$ & $\checkmark$ & $\checkmark$ \\
No proxy datasets
& $\checkmark$ & $\checkmark$ & $\times$ & $\times$ & $\checkmark$ & $\checkmark$ & $\checkmark$ \\
Dynamic FR detection
& $\checkmark$ & $\checkmark$ & $\checkmark$ & $\times$ & $\times$ & $\times$ & $\checkmark$ \\
\bottomrule
\end{tabularx}
\end{table*}

\subsection{WEF-Matrix}\label{WEF-matrix}
%long version
The existing method, known as WEF-defense, detects free-riders by using the \emph{Weight Evolving Frequency-matrix} (WEF-matrix) during local training \cite{WEF-defense}.
The WEF-matrix is a matrix that represents which parts of the penultimate layer of the local model were significantly updated during each local iteration by each client.
%The construction of the WEF-matrix $\mcF^{(T,t)}_{i}$ is as follows.
%We provide how to define the WEF-matrix for each client in the paragraph below. 
We define the WEF-matrix for each client below.

%Let $i$ denotes the $i$-th client,
Let $i$ denote the index of a given client, 
$T$ the current global communication round, and $t$ the current local iteration round.
The weight matrix of the penultimate layer for client $i$ at round $(T,t)$ is denoted by $w^{(T,t)}_{i}$, and let its size be $H \times W$.
First, at the beginning of the global communication round $T$, before starting local iterations, client $i$ initializes the WEF-matrix by
$\mathcal{F}_{i}^{(T,t=0)} = \zeros(H,W)$,
where $\zeros(H,W)$ denotes the zero matrix of size $H \times W$.
Next, after completing the $t$-th local iteration, the client calculates the threshold $\alpha_{i}^{(T,t)}$ for determining weight evolving frequency for each local iteration, which is defined by ${\alpha_{i}}^{(T,t)} =\frac{\sum_{j=1}^{H}\sum_{k=1}^{W}|w^{(T,t)}_{i,j,k} - w^{(T,t-1)}_{i,j,k} |}{H\times W}$,
where $w_{i,j,k}^{T,t}$ denotes the weight at the $j$-th row and $k$-th column of the penultimate layer after the $t$-th local iteration.
This threshold $\alpha_{i}^{(T,t)}$ is nothing but the average magnitude of weight changes before and after the local update.
Using this dynamic threshold, client $i$ updates the WEF-matrix as follows:
\begin{equation}
    \mathcal{F}^{(T,t)}_{i,j,k}=
    \begin{cases}
    \mathcal{F}^{(T,t-1)}_{i,j,k} + 1, & \text{if } \left| w^{(T,t)}_{i,j,k} - w^{(T,t-1)}_{i,j,k} \right| > \alpha_{i}^{(T,t)}, \\
    \mathcal{F}^{(T,t-1)}_{i,j,k}, & \text{otherwise}.
    \end{cases}\label{eq:WEFmatrix}
\end{equation}
That is, if the weight change in a specific part of the penultimate layer during local training is sufficiently large, which means it is greater than the dynamic threshold, then the corresponding element in the WEF-matrix is incremented.
Therefore, each element of the WEF-matrix $\mcF^{(T,t)}_{i}$ is a non-negative integer, and its maximum possible value is equal to the total number of local iterations.

For free-rider attacks that fabricate only the final model
parameters without performing local iterations, the attacker must also
construct a counterfeit WEF-matrix.
Let $w_i^{f,T}$ denote the fabricated parameters of client $i$ at global
communication round $T$.
We first compute the mean parameter change from the received global model as
\[
\alpha_i^{f,T}
=
\frac{1}{HW}
\sum_{j=1}^{H}\sum_{k=1}^{W}
\left|w_{i,j,k}^{f,T}-w_{g,j,k}^{T}\right|.
\]
The counterfeit WEF-matrix is then constructed as
\begin{equation}
\mcF^{f}_{j,k}
=
\begin{cases}
e, &
\left|w_{i,j,k}^{f,T}-w_{g,j,k}^{T}\right|>\alpha_i^{f,T},\\
0, & \text{otherwise},
\end{cases}
\label{eq:counterfeit_wef}
\end{equation}
where $e$ is the number of local iterations.
Thus, unlike a benign WEF-matrix generated over multiple local iterations,
this counterfeit WEF-matrix typically consists only of $0$ and $e$ while
reflecting the parameters that changed substantially.

\iffalse
%short version
WEF-defense detects free-riders using the \emph{Weight Evolving
Frequency-matrix} (WEF-matrix), which records significant parameter changes
in the penultimate layer during local training \cite{WEF-defense}.
Let $w_i^{(T,t)}\in\mathbb{R}^{H\times W}$ denote the penultimate-layer
weights of client $i$ at global communication round $T$ and local iteration
$t$.
Starting from $\mcF_i^{(T,0)}=\zeros(H,W)$, the client computes
\[
\alpha_i^{(T,t)}
=
\frac{1}{HW}
\sum_{j=1}^{H}\sum_{k=1}^{W}
\left|
w_{i,j,k}^{(T,t)}-w_{i,j,k}^{(T,t-1)}
\right|
\]
and updates
\begin{equation}
\mcF_{i,j,k}^{(T,t)}
=
\begin{cases}
\mcF_{i,j,k}^{(T,t-1)}+1,
&
\left|
w_{i,j,k}^{(T,t)}-w_{i,j,k}^{(T,t-1)}
\right|>\alpha_i^{(T,t)},\\
\mcF_{i,j,k}^{(T,t-1)}, & \text{otherwise}.
\end{cases}
\label{eq:WEFmatrix}
\end{equation}
Thus, each element counts how often the corresponding weight changes by
more than the mean absolute change over local iterations.

A free-rider that fabricates only the final parameters $w_i^{f,T}$ must also
construct a counterfeit WEF-matrix.
Using
$\alpha_i^{f,T}
=\frac{1}{HW}\sum_{j,k}|w_{i,j,k}^{f,T}-w_{g,j,k}^{T}|$,
we construct it as
\begin{equation}
\mcF_{j,k}^{f}
=
\begin{cases}
e,
&
\left|w_{i,j,k}^{f,T}-w_{g,j,k}^{T}\right|>\alpha_i^{f,T},\\
0, & \text{otherwise},
\end{cases}
\label{eq:counterfeit_wef}
\end{equation}
where $e$ is the number of local iterations.
Hence, this counterfeit WEF-matrix typically consists only of $0$ and $e$.
\fi
\subsection{WEF-Defense}\label{sec:wef-defense}
\iffalse
%short version
WEF-defense \cite{WEF-defense} detects free-riders from WEF-matrices
accumulated across global communication rounds:
$
\widetilde{\mcF}_i^{(T,t)}
=
\sum_{T'=1}^{T}\mcF_i^{(T',t)}$.
For each client, it computes the average Euclidean distance $\Dis_i$,
cosine similarity $\Cos_i$, and matrix value $\Avg_i$, and defines the
deviation score
\begin{equation}
    \begin{aligned}
        \Dev_i 
        ={}& 
        \frac{|\Dis_i-\overline{\Dis}|} {\sum_{j=1}^{N}|\Dis_j-\overline{\Dis}|}
        + \frac{|\Cos_i-\overline{\Cos}|} {\sum_{j=1}^{N}|\Cos_j-\overline{\Cos}|}
        \\
        &+ \frac{|\Avg_i-\overline{\Avg}|} {\sum_{j=1}^{N}|\Avg_j-\overline{\Avg}|}.
    \end{aligned}
    \label{eq:Dev}
\end{equation}
where $\Dis_i$ and $\Cos_i$ are the average pairwise Euclidean distance
and cosine similarity to the other clients, respectively, and $\Avg_i$ is
the mean WEF value.
$\overline{\Dis}$, $\overline{\Cos}$, and $\overline{\Avg}$ denote the
corresponding client-wise means.
Client $i$ is classified as a free-rider when
$\Dev_i>\xi$, where $\xi=\max_j\{\Dev_j\}-\epsilon$.
\fi
%long verion
This section describes WEF-defense \cite{WEF-defense}, an existing detection method that utilizes the WEF-matrix described above.

In WEF-defense, free-riders are detected in each global communication round using the WEF-matrix uploaded by each client.
In global communication round $T$, the central server calculates the accumulated WEF-matrix for each client $i$ as $\widetilde{\mcF}^{(T,t)}_{i} = \sum_{T'=1}^{T} \mcF^{(T',t)}_{i}$.
Then, the server computes the Euclidean distance, cosine similarity, and average value of $\widetilde{\mcF}^{(T,t)}_{i}$ for each client $i$.
Based on these three metrics, the deviation score $\Dev_{i}$ for client $i$ is calculated by
\begin{equation}
    \begin{aligned}
        \Dev_i 
        ={}& 
        \frac{|\Dis_i-\overline{\Dis}|} {\sum_{j=1}^{N}|\Dis_j-\overline{\Dis}|}
        + \frac{|\Cos_i-\overline{\Cos}|} {\sum_{j=1}^{N}|\Cos_j-\overline{\Cos}|}
        \\
        &+ \frac{|\Avg_i-\overline{\Avg}|} {\sum_{j=1}^{N}|\Avg_j-\overline{\Avg}|}.
    \end{aligned}
    \label{eq:Dev}
\end{equation}
where 
$\Dis_{i}$ is the average Euclidean distance between the WEF-matrix uploaded by $i$-th client and those of other clients, 
$\Cos_{i}$ is the average cosine similarity between the WEF-matrix of $i$-th client and those of other clients, and
$\Avg_{i}$ is the average value of the WEF-matrix of $i$-th client.
$\overline{\Dis}$, $\overline{\Cos}$, $\overline{\Avg}$ denote the mean values of $\Dis_{i}$, $\Cos_{i}$, $\Avg_{i}$, respectively.
Then the central server determines that $i$-th client is a free-rider if the calculated $\Dev_{i}$ exceeds the threshold $\xi = \max\{ \Dev_{i} \} - \epsilon$, where $\epsilon$ is a hyperparameter.

\section{Dynamic free-rider attack on WEF-defense}\label{sec:SFRonWEF}
\subsection{Limitations of WEF-defense under Dynamic Free-Riders}
\label{subsec:wef_limitations}
WEF-defense was originally designed under the assumption of static free-riders who never perform local training and remain free-riders throughout training.
Its detection mechanism relies on accumulating WEF-matrices across rounds, which improves separability when clients' behavior is consistent.
%However, for \emph{dynamic} free-riders who behave honestly in early rounds and then switch to free-riding, the accumulated WEF can remain similar to benign behavior for a long period, making dynamic detection difficult.
However, for \emph{dynamic} free-riders who behave honestly in early rounds and then switch to free-riding, the accumulated WEF can continue to reflect their earlier benign behavior for some time, making dynamic detection difficult.

According to our experiments, clients who behave honestly only during the first two global communication rounds and then switch to free-riding are extremely difficult to detect.
The left column of Tab.~\ref{SFRonWEFdefense} shows F1-score when using MNIST \cite{mnist}, Adult \cite{adult}, CIFAR-10 \cite{cifar10} and CIFAR-100 \cite{cifar10} as datasets. 
In this experiment, 30\% of clients become free-riders starting from the third global communication round. 
See Sec.~\ref{sec:Exp} for more details on experimental settings. 
Intuitively, it may be expected that detection becomes possible in the later rounds, where the WEF-matrix from free-riders has accumulated.
However, especially in the case of MNIST, many attacks remained undetected until the final round.

A simple solution to this issue is to detect free-riders using only the WEF-matrix of the current round, i.e. , using $\mcF^{(T,e)}_{i}$ instead of the accumulated $\widetilde{\mcF}^{(T,e)}_{i}$.
The results of this approach are shown in the right column of Tab.~\ref{SFRonWEFdefense}.
This method improved the F1-score in many cases, but still showed low F1-scores in most DWA settings.

\begin{table}[t]
\caption{Detection F1-score of WEF-Defense against existing dynamic free-rider attacks.}
\label{SFRonWEFdefense}
\centering
\scriptsize
\setlength{\tabcolsep}{2.6pt}
\renewcommand{\arraystretch}{0.90}
\begin{tabular}{@{}llccc@{}}
\toprule
Dataset & Dist. & Attack & WEF-Defense & WEF-na \\
\midrule
\multirow{8}{*}{MNIST} & \multirow{4}{*}{IID} & RWA & $0.406\!\pm\!0.514$ & $0.598\!\pm\!0.493$ \\
 &  & SPA & $0.306\!\pm\!0.419$ & $0.927\!\pm\!0.091$ \\
 &  & DWA & $0.198\!\pm\!0.271$ & $0.198\!\pm\!0.271$ \\
 &  & ADWA & $0.198\!\pm\!0.271$ & $0.927\!\pm\!0.081$ \\
\cmidrule(lr){2-5}
 & \multirow{4}{*}{Non-IID} & RWA & $0.194\!\pm\!0.435$ & $0.196\!\pm\!0.269$ \\
 &  & SPA & $0.166\!\pm\!0.365$ & $0.990\!\pm\!0.005$ \\
 &  & DWA & $0.197\!\pm\!0.270$ & $0.194\!\pm\!0.266$ \\
 &  & ADWA & $0.197\!\pm\!0.270$ & $0.778\!\pm\!0.260$ \\
\midrule
\multirow{8}{*}{Adult} & \multirow{4}{*}{IID} & RWA & $0.976\!\pm\!0.006$ & $0.986\!\pm\!0.011$ \\
 &  & SPA & $0.907\!\pm\!0.016$ & $0.988\!\pm\!0.007$ \\
 &  & DWA & $0.198\!\pm\!0.271$ & $0.198\!\pm\!0.271$ \\
 &  & ADWA & $0.689\!\pm\!0.261$ & $0.988\!\pm\!0.006$ \\
\cmidrule(lr){2-5}
 & \multirow{4}{*}{Non-IID} & RWA & $0.965\!\pm\!0.014$ & $0.984\!\pm\!0.003$ \\
 &  & SPA & $0.913\!\pm\!0.008$ & $0.989\!\pm\!0.006$ \\
 &  & DWA & $0.395\!\pm\!0.221$ & $0.396\!\pm\!0.221$ \\
 &  & ADWA & $0.492\!\pm\!0.005$ & $0.971\!\pm\!0.013$ \\
\midrule
\multirow{8}{*}{CIFAR-10} & \multirow{4}{*}{IID} & RWA & $0.875\!\pm\!0.110$ & $0.895\!\pm\!0.223$ \\
 &  & SPA & $0.920\!\pm\!0.018$ & $0.995\!\pm\!0.001$ \\
 &  & DWA & $0.554\!\pm\!0.130$ & $0.698\!\pm\!0.272$ \\
 &  & ADWA & $0.561\!\pm\!0.130$ & $0.722\!\pm\!0.218$ \\
\cmidrule(lr){2-5}
 & \multirow{4}{*}{Non-IID} & RWA & $0.749\!\pm\!0.426$ & $0.894\!\pm\!0.224$ \\
 &  & SPA & $0.750\!\pm\!0.420$ & $0.994\!\pm\!0.002$ \\
 &  & DWA & $0.198\!\pm\!0.271$ & $0.099\!\pm\!0.221$ \\
 &  & ADWA & $0.202\!\pm\!0.263$ & $0.954\!\pm\!0.061$ \\
\midrule
\multirow{8}{*}{CIFAR-100} & \multirow{4}{*}{IID} & RWA & $0.986\!\pm\!0.005$ & $0.000\!\pm\!0.000$ \\
 &  & SPA & $0.996\!\pm\!0.001$ & $0.902\!\pm\!0.190$ \\
 &  & DWA & $0.780\!\pm\!0.196$ & $0.522\!\pm\!0.034$ \\
 &  & ADWA & $0.788\!\pm\!0.202$ & $0.688\!\pm\!0.040$ \\
\cmidrule(lr){2-5}
 & \multirow{4}{*}{Non-IID} & RWA & $0.207\!\pm\!0.284$ & $0.299\!\pm\!0.446$ \\
 &  & SPA & $0.189\!\pm\!0.416$ & $0.911\!\pm\!0.191$ \\
 &  & DWA & $0.101\!\pm\!0.222$ & $0.294\!\pm\!0.437$ \\
 &  & ADWA & $0.099\!\pm\!0.222$ & $0.349\!\pm\!0.325$ \\
\bottomrule
\end{tabular}
\vspace{1pt}
\begin{minipage}{\columnwidth}
\scriptsize
Values are mean $\pm$ sample standard deviation over five random seeds.
\textit{Abbreviations:} WEF-na, WEF-Defense with no accumulation of the WEF matrix.
\end{minipage}
\end{table}

\subsection{Adaptive WEF-Camouflage Attack}
\label{subsec:awca}
Based on our experimental results, DWA can be considered an effective attack against WEF-defense.
However, this attack has the following two limitations:
(i) When multiple free-riders exist, they generate identical fake weight parameters.
(ii) The generated counterfeit WEF-matrix becomes a binary matrix consisting only of values $0$ and $e$.
These characteristics may allow for ad-hoc detection of DWA.
To address this, we propose a more powerful free-rider attack, called adaptive WEF-camouflage attack (AWCA).
This attack is an extension of ADWA, and it generates counterfeit parameters sequentially at each local iteration..
While existing attacks compute the WEF-matrix using the global model before training and a single counterfeit model after local training,
our proposed attack generates counterfeit weights sequentially at each local iteration.
This attack enables the attacker to counterfeit the WEF-matrix in a manner more similar to benign clients.
In this attack, the local model weight $w_i^{(T,t)}$ of the $i$-th client at the $t$-th local iteration in the $T$-th global communication round is generated by
\begin{equation}
    w_{i}^{(T,t)} = w_{i}^{(T,t-1)} + \frac{w_{g}^{T}-w_{g}^{T-1}}{e} + N(0,\sigma)\label{eq:AWCA}.
\end{equation}
In the above equation, we set $w_{i}^{(T,0)} = w_{g}^{T}$ for convenience, where $w_g^T$ denotes the global model received from the central server at global communication round $T$. 
We also note that $e$ denotes the total number of local iterations and $N(0,\sigma)$ denotes the Gaussian noise.
%By generating fake model parameters for each local iteration in this manner,
%the counterfeited WEF-matrix can be constructed in the same way as described in Sec.~\ref{WEF-matrix}.
%Algorithm~\ref{alg:AWCA} summarizes the procedure of the proposed attack.
The overall procedure for generating the sequential counterfeit weights and the corresponding WEF-matrix is summarized in Algorithm~\ref{alg:AWCA}.

Regarding the standard deviation $\sigma$ of the Gaussian noise,
we found through simple experiments that small noise values are effective: $\sigma = 10^{-5}$ for image datasets such as MNIST, CIFAR-10 and CIFAR-100, and $\sigma = 10^{-6}$ for tabular datasets such as Adult.
Tab.~\ref{tab:wef_novel_dynamic_attack} shows the F1-score of WEF-defense against this attack.
Like DWA, this proposed attack was difficult to detect using WEF-defense.

In summary, the existing method WEF-defense struggles to dynamically detect clients who change into free-riders during training. 
New detection methods are required to improve detection performance, especially for DWA and our proposed AWCA.

\begin{algorithm}[t]
\caption{Adaptive WEF-camouflage attack (AWCA)}
\label{alg:AWCA}
\footnotesize
\begin{algorithmic}[1]
\STATE \textbf{Input:} broadcast global models $w_g^{T}, w_g^{T-1}$; local iterations $e$; noise std.\ $\sigma$
\STATE \textbf{Output:} counterfeit weights $w_i^{(T,e)}$ and WEF-matrix $\mathcal{F}_i^{(T,e)}$
\STATE Initialize $w_i^{(T,0)} \gets w_g^{T}$ and $\mathcal{F}_i^{(T,0)} \gets \mathbf{0}$ \COMMENT{penultimate layer of size $H \times W$}
\STATE $\Delta w \gets (w_g^{T}-w_g^{T-1})/e$
\FOR{$t=1$ to $e$}
  \STATE $w_i^{(T,t)} \gets w_i^{(T,t-1)} + \Delta w + N(0,\sigma)$
  \STATE $\alpha_i^{(T,t)} \gets \frac{1}{HW}\sum_{j=1}^{H}\sum_{k=1}^{W}\left|w_{i,j,k}^{(T,t)} - w_{i,j,k}^{(T,t-1)}\right|$
  \STATE Update $\mathcal{F}_i^{(T,t)}$ element-wise:
  \STATE \hspace{\algorithmicindent} $\mathcal{F}_{i,j,k}^{(T,t)} \gets \mathcal{F}_{i,j,k}^{(T,t-1)} + 1$ if $\left|w_{i,j,k}^{(T,t)} - w_{i,j,k}^{(T,t-1)}\right| > \alpha_i^{(T,t)}$, else $\mathcal{F}_{i,j,k}^{(T,t)} \gets \mathcal{F}_{i,j,k}^{(T,t-1)}$.
\ENDFOR
\RETURN $w_i^{(T,e)}, \mathcal{F}_i^{(T,e)}$
\end{algorithmic}
\end{algorithm}

\begin{table}[t]
\caption{Detection F1-score of WEF-Defense against the novel dynamic free-rider attack.}
\label{tab:wef_novel_dynamic_attack}
\centering
\scriptsize
\setlength{\tabcolsep}{3.2pt}
\renewcommand{\arraystretch}{0.95}
\begin{tabular}{@{}llcc@{}}
\toprule
Dataset & Dist. & WEF-Defense & WEF-na \\
\midrule
\multirow{2}{*}{MNIST} & IID & $0.198\!\pm\!0.271$ & $0.198\!\pm\!0.271$ \\
 & Non-IID & $0.197\!\pm\!0.270$ & $0.194\!\pm\!0.266$ \\
\midrule
\multirow{2}{*}{Adult} & IID & $0.198\!\pm\!0.271$ & $0.198\!\pm\!0.271$ \\
 & Non-IID & $0.395\!\pm\!0.221$ & $0.396\!\pm\!0.221$ \\
\midrule
\multirow{2}{*}{CIFAR-10} & IID & $0.554\!\pm\!0.130$ & $0.698\!\pm\!0.272$ \\
 & Non-IID & $0.198\!\pm\!0.271$ & $0.099\!\pm\!0.221$ \\
\midrule
\multirow{2}{*}{CIFAR-100} & IID & $0.781\!\pm\!0.197$ & $0.436\!\pm\!0.133$ \\
 & Non-IID & $0.199\!\pm\!0.272$ & $0.294\!\pm\!0.437$ \\
\bottomrule
\end{tabular}
\vspace{1pt}
\begin{minipage}{\columnwidth}
\scriptsize
Values are mean $\pm$ sample standard deviation over five random seeds.
\textit{Abbreviations:} WEF-na, WEF-Defense with no accumulation of the WEF matrix.
\end{minipage}
\end{table}

\section{S2-WEF}\label{sec:S2WEF}
\subsection{Overview}
As discussed in Sec.~\ref{sec:SFRonWEF}, dynamic free-riders are difficult to detect with WEF-defense under global-model-mimicking attacks such as DWA and AWCA.
These attacks generate fake updates that are identical or nearly identical to the differences between previously broadcast global models.
Motivated by this observation, we detect clients whose submitted WEF-matrices resemble WEF patterns simulated on the server from past global model differences, which the server can readily record.

This approach specializes in detecting attacks that mimic the global model. 
In contrast, we expect the conventional WEF-defense to be effective against other types of attacks (e.g., RWA and SPA). 
In other words, these attacks can be detected by comparing the WEF-matrices submitted by clients with one another.

\begin{figure*}
    \begin{center}
        \includegraphics[width=\linewidth]{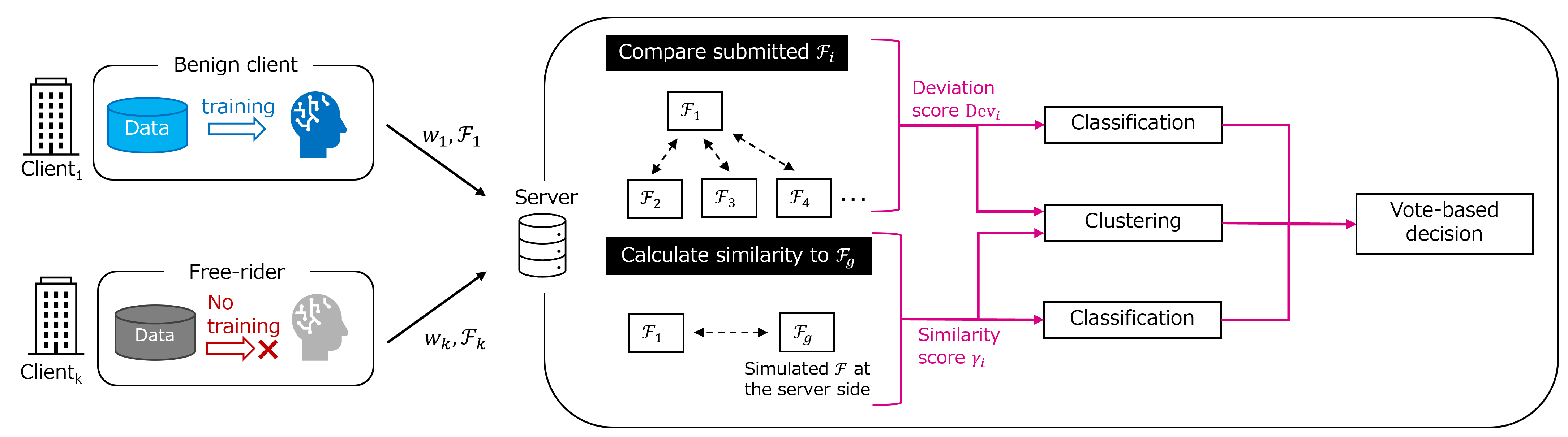}
        \caption{Overview of S2-WEF}
        \label{OverviewS2-WEF}
    \end{center}
\end{figure*}

However, the central server does not know in advance which type of attack it will face. 
Naively combining two detection mechanisms may increase false positives. 
Therefore, as shown in Fig.~\ref{OverviewS2-WEF}, we perform (i) two-dimensional clustering based on both the similarity score to the simulated WEF-matrix and the deviation score derived from pairwise comparisons of submitted WEF-matrices and (ii) threshold-based classification for each score. 
Moreover, for the suspicious client group identified by clustering, we conduct (iii) a vote-based decision using the outcomes of the threshold-based classifications. 
Specifically, we label the group as free-riders only when at least half of the clients in the cluster exceed the threshold of either score.
This design enables us to detect diverse attacks while suppressing false positives.
%The detailed implementation of our method is presented in Algorithm~\ref{alg:S2WEF}.
Algorithm~\ref{alg:S2WEF} summarizes the procedure.
\iffalse
\fi

\subsection{Simulation of Free-Riders on the Server Side}
\iffalse
\fi
At each global communication round $T$, the central server initializes the simulated WEF-matrix by
$
     {\mathcal{F}_{g}}^{(T,0)} = \zeros(H,W),
$
where $H\times W$ corresponds to the size of the penultimate layer of the global model. 
Next, using the two global models previously broadcast to clients in past rounds, the server computes the WEF-matrix by
\begin{equation}
    \mathcal{F}^{(T,1)}_{g,j,k}=
\begin{cases}
\mathcal{F}^{(T,0)}_{g,j,k} + 1, & \text{if } \left| w^{T}_{g,j,k} - w^{T-1}_{g,j,k} \right| > \alpha_{g}^{T}, \\
\mathcal{F}^{(T,0)}_{g,j,k}, & \text{otherwise}.
\end{cases}
\end{equation}
The dynamic threshold ${\alpha_{g}}^{T}$ is calculated as
${\alpha_{g}}^{T} =\frac{\sum_{j=1}^{H}\sum_{k=1}^{W}|w^{T}_{g,j,k} - w^{T-1}_{g,j,k} |}{H\times W}$.
\iffalse
To match the maximum value of the WEF-matrix uploaded by each client, the server multiplies the computed $\mathcal{F}^{(T,1)}_{g,j,k}$ by the number of local epochs $e$: $\mathcal{F}^{(T,e)}_{g,j,k} =e\cdot\mathcal{F}^{(T,1)}_{g,j,k}\label{eq:F-g}$.
\fi
To match the maximum value of the WEF-matrix uploaded by each client, the server multiplies the computed $\mathcal{F}^{(T,1)}_{g,j,k}$ by the number of local iterations $e$ and obtains 
\begin{equation}
\mathcal{F}^{(T,e)}_{g,j,k} =e\cdot\mathcal{F}^{(T,1)}_{g,j,k}\label{eq:F-g}.
\end{equation}
This generated WEF-matrix $\mathcal{F}^{(T,e)}_{g,j,k}$ equals what a free-rider performing DWA would generate.
Using this simulated WEF-matrix as the basis, the server can defend against attacks that mimic the global model.

Note that in the case of ADWA and AWCA, increasing the standard deviation $\sigma$ of the Gaussian noise results in a WEF-matrix that becomes dissimilar to the simulated one. 
In such cases, the defense relies on the deviation score $\text{Dev}_{i}$ (see (\ref{eq:Dev})), which is noise-sensitive.

\subsection{Similarity to the Simulated WEF-Matrix}\label{sec:compute-gamma}
The central server calculates the similarity between the simulated WEF-matrix
$\mathcal{F}_{g}$, computed in (\ref{eq:F-g}), and the WEF-matrix
$\mathcal{F}_{i}$ submitted by each client.
We first define their cosine similarity as
\begin{equation}
    \Cos_{i,g}
    =
    \frac{
        \mathcal{F}_{i} \cdot \mathcal{F}_{g}
    }{
        \|\mathcal{F}_{i}\|_{2}
        \|\mathcal{F}_{g}\|_{2}
    },
\end{equation}
where $\cdot$ denotes the matrix inner product and
$\|\cdot\|_{2}$ denotes the Euclidean $L^{2}$-norm of the vectorized
matrix.
We further compute the $L^{2}$-distance between the two WEF-matrices as
\begin{equation}
    \|\mathcal{F}_{i}-\mathcal{F}_{g}\|_{2}
    =
    \sqrt{
        \sum_{j=1}^{H}
        \sum_{k=1}^{W}
        \left(
            \mathcal{F}_{i,j,k}
            -
            \mathcal{F}_{g,j,k}
        \right)^{2}
    },
\end{equation}
where $H$ and $W$ denote the number of rows and columns, respectively.
Note that, unlike the original WEF-defense, both
$\mathcal{F}_{g}$ and $\mathcal{F}_{i}$ are computed independently
at each global communication round without accumulation.

Using the cosine similarity and the $L^{2}$-distance, the central server
computes the similarity score $\gamma_i$ as
\begin{equation}
    \gamma_{i}
    =
    \frac{
        \Cos_{i,g}
    }{
        \|\mathcal{F}_{i}-\mathcal{F}_{g}\|_{2}
        + \varepsilon_{\gamma}
    },
    \label{eq:gamma-i}
\end{equation}
where $\varepsilon_{\gamma}>0$ is a small constant introduced to avoid
division by zero when the submitted and simulated WEF-matrices are
identical.
The score $\gamma_i$ becomes large when $\mathcal{F}_{i}$ and
$\mathcal{F}_{g}$ exhibit both a similar update pattern and a small
absolute discrepancy. Specifically, cosine similarity captures the
directional similarity between the two WEF patterns, whereas the
$L^{2}$-distance captures their element-wise discrepancy.
Combining these two measures therefore emphasizes clients whose submitted
WEF-matrices closely resemble the WEF pattern expected from
global-model-mimicking attacks.

%Compared with the $L^{1}$-distance, the $L^{2}$-distance is less affected
%by the accumulation of many small element-wise discrepancies while
%placing relatively greater emphasis on larger deviations.
%This property is useful for distinguishing structurally similar WEF
%patterns even when small variations are distributed across multiple
%elements.

\subsection{Clustering and Classification}\label{sec:clustering_classification}
At each global communication round, the server computes two
anomaly-related scores for each client $i$: the similarity score
$\gamma_i$ (Sec.~\ref{sec:compute-gamma}) and the deviation score
$\mathrm{Dev}_i$ (Sec.~\ref{sec:wef-defense}).
We jointly use these scores in two ways.
First, we perform clustering in the two-dimensional score space to identify
a group of potentially suspicious clients.
Second, we independently apply threshold-based classification to each score.
The clustering and classification results are subsequently combined by the
vote-based procedure described in Sec.~\ref{sec:majority_vote}.

Since our threat model assumes that free-riders constitute less than half of
the clients, the median of the client scores is expected to remain anchored
to benign behavior.
We therefore use median-based robust statistics both for score
standardization and for defining the threshold of $\gamma_i$.

\paragraph{Agglomerative hierarchical clustering in a robustly standardized
2D space.}

To jointly analyze $\gamma_i$ and $\mathrm{Dev}_i$, we first map the two
scores into a common scale using robust standardization based on the median
and the median absolute deviation (MAD).
For $x \in \{\gamma,\mathrm{Dev}\}$, we compute
\begin{equation}
z_i^{(x)}
=
\frac{
x_i-\operatorname{median}(\{x_j\}_{j=1}^{N})
}{
\operatorname{MAD}(\{x_j\}_{j=1}^{N})+\varepsilon_z
},
\label{eq:robust_z}
\end{equation}
where
$
\operatorname{MAD}(\{x_j\}_{j=1}^{N})
=
\operatorname{median}
\left(
\left\{
\left|
x_j-\operatorname{median}(\{x_k\}_{k=1}^{N})
\right|
\right\}_{j=1}^{N}
\right),
$
and $\varepsilon_z>0$ is a small constant that prevents numerical
instability when the MAD is zero or close to zero.
Each client is then represented by
\begin{equation}
\mathbf{z}_i
=
\left(
z_i^{(\gamma)},
z_i^{(\mathrm{Dev})}
\right),
\end{equation}
and we apply agglomerative hierarchical clustering to
$\{\mathbf{z}_i\}_{i=1}^{N}$ using Ward's linkage criterion and the
Euclidean distance metric \cite{ward-hierarchical}.

We adopt hierarchical clustering for two practical reasons.
First, unlike K-means, it does not depend on random centroid
initialization and therefore provides deterministic and reproducible
partitions.
Second, our clustering procedure can conservatively fall back to a
single-cluster solution when the tentative partition does not exhibit
sufficient separation, which helps suppress false positives in rounds
without clear anomalous behavior.

Specifically, we first obtain a tentative two-cluster partition and compute
its silhouette coefficient $S_2$.
If $S_2 < 0.40$,
we regard the separation as insufficient and select $K=1$.
In this case, no suspicious cluster is produced.
Otherwise, we retain the two-cluster solution ($K=2$).
When $K=2$ is selected, we compute the centroid of each cluster in the
robustly standardized $\mathbf{z}$-space and designate the cluster whose
centroid has the larger Euclidean distance from the origin as the
\emph{suspicious cluster}.
Since the origin corresponds to the median-centered behavior of the
clients, this rule selects the group that deviates more strongly from the
benign-majority reference.

\paragraph{Threshold-based classification for each score.}

In parallel with clustering, we perform threshold-based classification
using the original, non-standardized scores.

For the similarity score, we define an adaptive threshold based on the
median and MAD of the scores in the current round:

\begin{equation}
\tau_{\gamma}
=
\operatorname{median}(\{\gamma_j\}_{j=1}^{N})
+
k_{\gamma}
\operatorname{MAD}(\{\gamma_j\}_{j=1}^{N}).
\label{eq:gamma-thr-value}
\end{equation}

Client $i$ is flagged by the similarity score if
$
\gamma_i > \tau_{\gamma}.
%\label{eq:gamma_thr}
$
This threshold adapts to the round-wise dispersion of $\gamma$ and is
therefore less dependent on the absolute scale of the similarity score than
a fixed multiplicative threshold.
We set $k_{\gamma}=5$ based on a sensitivity analysis; the effect of
$k_{\gamma}$ on detection performance is evaluated in
Sec.~\ref{sec:gamma-sensitivity}.

For the deviation score, we follow the thresholding rule introduced by
WEF-defense \cite{WEF-defense}.
The original WEF-defense uses a margin of $0.05$ in experiments with
$N=10$ clients.
Because $\mathrm{Dev}_i$ is normalized using statistics aggregated over
all $N$ clients, we scale this margin inversely with the number of clients
to accommodate different client populations.
We therefore define

\begin{equation}
\tau_{\mathrm{Dev}}
=
\max(\{\mathrm{Dev}_j\}_{j=1}^{N})
-
0.05\left(\frac{10}{N}\right),
\label{eq:dev-thr-value}
\end{equation}

and flag client $i$ if
$ 
\mathrm{Dev}_i > \tau_{\mathrm{Dev}}.
%\label{eq:dev_thr}
$
%For $N=10$, this definition reduces to the original threshold used in
%WEF-defense.
%The resulting binary classifications based on $\gamma_i$ and
%$\mathrm{Dev}_i$ are combined with the suspicious cluster identified above
%through the majority-vote decision described in
%Sec.~\ref{sec:majority_vote}.
\iffalse

\fi
\subsection{Vote-Based Decision}
\label{sec:majority_vote}
The suspicious cluster identified in
Sec.~\ref{sec:clustering_classification} is verified using the
score-wise threshold tests before assigning free-rider labels.
If clustering selects $K=1$, no free-riders are detected in round $T$.
Otherwise, let $\mathcal{C}_{\mathrm{sus}}(T)$ denote the suspicious
cluster. We compute the proportions of its clients exceeding the
$\gamma$ and $\mathrm{Dev}$ thresholds as
\begin{equation}
\begin{aligned}
p_\gamma(T)
&=
\frac{1}{|\mathcal{C}_{\mathrm{sus}}(T)|}
\sum_{i\in\mathcal{C}_{\mathrm{sus}}(T)}
\mathbf{1}\!\left[\gamma_i(T)>\tau_\gamma(T)\right],\\
p_{\mathrm{Dev}}(T)
&=
\frac{1}{|\mathcal{C}_{\mathrm{sus}}(T)|}
\sum_{i\in\mathcal{C}_{\mathrm{sus}}(T)}
\mathbf{1}\!\left[\mathrm{Dev}_i(T)>\tau_{\mathrm{Dev}}(T)\right].
\end{aligned}
\label{eq:vote_rates}
\end{equation}

Free-riders are detected if at least half of the suspicious-cluster
clients exceed either threshold:
\begin{equation}
\mathrm{FreeRiderDetected}(T)
=
\mathbf{1}\!\left[
p_\gamma(T)\ge\frac{1}{2}
\;\lor\;
p_{\mathrm{Dev}}(T)\ge\frac{1}{2}
\right].
\label{eq:majority_rule}
\end{equation}
If $\mathrm{FreeRiderDetected}(T)=1$, all clients in
$\mathcal{C}_{\mathrm{sus}}(T)$ are labeled as free-riders for round $T$;
otherwise, all clients are treated as benign.
This verification prevents clustering alone from triggering free-rider
labels and thereby helps suppress false positives.

\subsection{Computational Complexity Analysis}\label{sec:complexity}
We analyze the computational cost of S2-WEF per global communication round.
Let $H \times W$ be the size of the penultimate layer, $e$ the total number of local iterations, and $N$ the number of clients.
On the client side, constructing the WEF-matrix scans $H \times W$ weights for each local iteration, yielding $O(e \cdot H \cdot W)$ per client.
On the server side, simulating the WEF-matrix from consecutive global models costs $O(H \cdot W)$, and computing $\{\gamma_i\}_{i=1}^{N}$ costs $O(N \cdot H \cdot W)$.
Computing $\mathrm{Dev}$ requires pairwise comparisons among clients' WEF-matrices, leading to $O(N^2 \cdot H \cdot W)$. 
In addition, threshold-based classification on raw $\gamma$ and $\mathrm{Dev}$ costs $O(N)$, while agglomerative hierarchical clustering and the associated validity checks incur $O(N^2)$. 

Therefore, the overall complexity per round is
$
O\!\left(N \cdot e \cdot H \cdot W \;+\; N^2 \cdot H \cdot W \;+\; N^2\right),
$
where $N^2 \cdot H \cdot W$ is the dominant term, which is also common to the existing method WEF-defense \cite{WEF-defense}. 
While this is acceptable for cross-silo FL with small $N$, it may be impractical for cross-device FL with large $N$.
%A common remedy is to approximate $\mathrm{Dev}$ by comparing each client only with a sampled subset of $M \ll N$ clients, reducing the $\mathrm{Dev}$-related cost to $O(N \cdot M \cdot H \cdot W)$.
One possible direction is to approximate $\mathrm{Dev}$ by comparing each client only with a sampled subset of $M \ll N$ clients, reducing the $\mathrm{Dev}$-related cost to $O(N \cdot M \cdot H \cdot W)$; in addition, the clustering step can be replaced by a more scalable alternative (e.g., density-based clustering such as HDBSCAN \cite{hdbscan,flame}) when $N$ is large.

\begin{algorithm}[t]
    \caption{S2-WEF Free-Rider Detection}
    \label{alg:S2WEF} 
    \footnotesize 
    \begin{algorithmic}[1] 
        \STATE 
        \textbf{Input:} total global communication rounds $E$; local iterations $e$; initial global model $w_g^0$ 
        \STATE \textbf{Output:} $w_g^{E}$, $\mathrm{FreeRiderList}$ 
        \STATE Initialize $\mathrm{FreeRiderList}[T] \gets \emptyset$ for all $T \in \{0,\dots,E-1\}$ \FOR{$T=0$ to $E-1$} 
        \STATE Clients upload $(\mathcal{F}_i^{(T,e)}, w_i^{(T,e)})$ (WEF by (\ref{eq:WEFmatrix}); AWCA: sequential, otherwise one-step scaling by (\ref{eq:counterfeit_wef})). 
        \STATE Server computes $\mathcal{F}_g^{(T,e)}$, $\{\Dev_i,\gamma_i\}_{i=1}^{N}$ using (\ref{eq:Dev}) and (\ref{eq:gamma-i}), threshold flags using (\ref{eq:gamma-thr-value}) and (\ref{eq:dev-thr-value}), and $\{\mathbf{z}_i^{(T)}\}_{i=1}^{N}$ using (\ref{eq:robust_z}). 
        \STATE Apply agglomerative hierarchical clustering to $\{\mathbf{z}_i^{(T)}\}_{i=1}^{N}$; obtain $K\in\{1,2\}$ and, if $K=2$, $\mathcal{C}_{\mathrm{sus}}(T)$. \IF{$K=1$} 
        \STATE $\mathrm{FreeRiderList}[T] \gets \emptyset$. 
        \ELSE 
        \STATE Compute $p_\gamma(T)$ and $p_{\mathrm{Dev}}(T)$ on $\mathcal{C}_{\mathrm{sus}}(T)$ using (\ref{eq:vote_rates}). 
        \STATE $\mathrm{FreeRiderList}[T] \gets \mathcal{C}_{\mathrm{sus}}(T)$ \textbf{if} $p_\gamma(T)\ge\tfrac{1}{2}$ \textbf{or} $p_{\mathrm{Dev}}(T)\ge\tfrac{1}{2}$; \textbf{otherwise}, $\mathrm{FreeRiderList}[T]\gets\emptyset$. \ENDIF \STATE $\mathcal{B}(T) \gets \{1,\dots,N\} \setminus \mathrm{FreeRiderList}[T]$. 
        \STATE Aggregate $w_g^{T+1} \gets \frac{1}{|\mathcal{B}(T)|} \sum_{i\in\mathcal{B}(T)} w_i^{(T,e)}$; broadcast $w_g^{T+1}$. 
        \ENDFOR 
        \RETURN $w_g^{E}$, $\mathrm{FreeRiderList}$ \end{algorithmic} \end{algorithm}
\section{Experiments}\label{sec:Exp}
We evaluate S2-WEF from five perspectives: detection performance against existing baselines, contribution of individual components, sensitivity to the similarity-score threshold, robustness to different attack settings, and scalability with the number of clients.
We further evaluate the impact of S2-WEF on the accuracy of the global model. Unless otherwise stated, experiments follow the setup of the main baseline comparison in Sec.~\ref{sec:main-experiment}.
\subsection{Comparison with Baseline Methods}
\label{sec:main-experiment}

\subsubsection{Experimental Setup}

\textbf{Datasets and models.}
We use four datasets: MNIST \cite{mnist}, Adult \cite{adult},
CIFAR-10 \cite{cifar10}, and CIFAR-100 \cite{cifar10}.
MNIST is a handwritten digit dataset, Adult is a tabular dataset for
binary income classification, and CIFAR-10 and CIFAR-100 are natural
image datasets with 10 and 100 classes, respectively.
We use LeNet \cite{LeNet} for MNIST, an MLP for Adult,
and ResNet-based architectures \cite{ResNet} for CIFAR-10 and CIFAR-100.
The main hyperparameters are summarized in
Tab.~\ref{tab:hyperparameters}.

\textbf{Data distribution and clients.}
We consider both IID and non-IID data distributions.
For the non-IID setting, client data are sampled according to a
Dirichlet distribution with concentration parameter $\beta=0.5$,
following WEF-defense \cite{WEF-defense}.
%As discussed in Sec.~\ref{sec:intro}, 
We primarily focus on cross-silo FL;
therefore, we set the number of clients to $N=10$.
%The free-rider ratio is fixed at $10\%$.
%The impact of varying the free-rider ratio is evaluated separately in Sec.~\ref{sec:attack-robustness}.

\textbf{Attacks and free-rider behavior.}
We evaluate five attacks: RWA, SPA, DWA, ADWA, and AWCA.
We set $R=10^{-3}$ for RWA, $\sigma=10^{-3}$ for ADWA, and
$\sigma=10^{-5}$ for AWCA on MNIST, CIFAR-10, and CIFAR-100 and
$\sigma=10^{-6}$ on Adult.
As in Sec.~\ref{sec:SFRonWEF}, all clients behave honestly during the
first two global communication rounds.
From the third round onward, $10\%$ of the clients switch to
free-riding and continue the corresponding attack until the end of
training.

\textbf{Baselines.}
We compare S2-WEF with STD-DAGMM \cite{STD-DAGMM} and WEF-defense
\cite{WEF-defense}.
We use STD-DAGMM as a representative DAGMM-based detector rather than
its variants FRAD \cite{FRAD} and ZeTFRi \cite{ZeTFRi}, which require
additional side information and a proxy dataset, respectively.
We also exclude contribution-evaluation methods because CFFL requires
a proxy dataset, while RFFL has been reported to be ineffective against
DWA \cite{WEF-defense}.
For WEF-defense, we disable cross-round WEF accumulation because
Sec.~\ref{sec:SFRonWEF} shows that accumulation degrades detection of
dynamic free-riders.

\textbf{Evaluation metrics.}
We evaluate detection performance using precision, recall, and F1-score.
For each run, we aggregate true positives, false positives, and false
negatives over all global communication rounds and calculate the three
metrics from the aggregated counts.
Each experimental setting is repeated with five different random seeds,
and we report the mean and standard deviation as
$\mathrm{mean}\pm\mathrm{std}$.

\begin{table*}[t]
\caption{Dataset and hyperparameter settings.}
\label{tab:hyperparameters}
\centering
\scriptsize
\setlength{\tabcolsep}{4.0pt}
\renewcommand{\arraystretch}{1.05}
\begin{tabular}{@{}lrcclccrr@{}}
\toprule
Dataset & Samples & Dimensions & Classes & Model & Learning rate & Momentum & Global rounds & Batch size \\
\midrule
MNIST     & 70,000 & $28 \times 28$ & 10  & LeNet  & 0.005  & 0.0001 & 50  & 32 \\
Adult     & 23,374 & 14              & 2   & MLP    & 0.0001 & 0.0001 & 50  & 32 \\
CIFAR-10  & 60,000 & $32 \times 32$ & 10  & ResNet & 0.01   & 0.9    & 80  & 32 \\
CIFAR-100 & 60,000 & $32 \times 32$ & 100 & ResNet & 0.01   & 0.9    & 100 & 32 \\
\bottomrule
\end{tabular}
\end{table*}

\subsubsection{Results and Analysis}
\label{sec:main-results}
%long version of the results and analysis
The results are shown in
Tabs.~\ref{tab:f1_main_mnist_adult} and
\ref{tab:f1_main_cifar}.
Overall, S2-WEF achieves consistently high detection performance across
datasets, data distributions, and attack types.
Among the 40 evaluated settings, S2-WEF achieves a mean F1-score of at
least $0.90$ in 39 settings, and its lowest mean F1-score is $0.861$.
Moreover, except for several settings on Adult under the non-IID
distribution, the standard deviations of S2-WEF are small, indicating
stable performance across different random seeds.

In contrast, the baseline methods show substantial performance
degradation in several settings.
STD-DAGMM is particularly unstable on MNIST and Adult, and its F1-score
becomes close to zero under some global-model-mimicking attacks.
WEF-na performs well against several attacks, but its performance can
also degrade substantially against DWA and AWCA.
For example, under the non-IID setting, WEF-na achieves an F1-score of
$0.184\pm0.174$ against DWA and $0.008\pm0.018$ against AWCA on MNIST,
and fails to detect both attacks on Adult with F1-score
$0.000\pm0.000$.
Similarly, on CIFAR-10 under the non-IID setting, its F1-score is
$0.000\pm0.000$ against both DWA and AWCA.
In contrast, S2-WEF maintains F1-scores above $0.9$ in these settings.
These results support the effectiveness of explicitly simulating the WEF
patterns of global-model-mimicking attacks on the server side.
At the same time, S2-WEF maintains high detection performance against
RWA, SPA, and ADWA, suggesting that combining the simulation-based
similarity score with the deviation score enables robust detection across
different free-rider strategies.

%S2-WEF also tends to exhibit particularly stable performance on
%CIFAR-10 and CIFAR-100.
%One possible explanation is that the ResNet-based models used for these
%datasets provide higher-dimensional WEF representations than the models
%used for MNIST and ADULT, potentially providing richer information for
%distinguishing benign clients from free-riders.
%However, this interpretation should be regarded as a possible factor
%rather than a definitive explanation, since both the datasets and model
%architectures differ across these experimental settings.

The most challenging setting for S2-WEF is Adult under the non-IID
distribution.
Although the mean F1-score remains above $0.9$ for four of the five
attacks, the variance across seeds is larger than in the other datasets.
The lowest performance is observed for SPA, with an F1-score of
$0.861\pm0.265$.
Inspection of the per-seed results showed that this degradation is
concentrated in one of the five seeds, whereas the remaining four seeds
exhibit stable detection performance.
In the degraded seed, the WEF-matrix of a benign client happened to
produce a large similarity score $\gamma$, indicating an unusually high
similarity to the WEF-matrix simulated on the server side.
This resulted in false-positive detections.
This observation reveals a limitation of the simulation-based similarity
score: under strong data heterogeneity, a benign WEF pattern may
occasionally resemble the simulated attack pattern.
Further suppressing such false positives is therefore an important
direction for improving S2-WEF.
Another possible factor behind the performance degradation on Adult is
the size of the WEF representation.
The MLP used for Adult provides a smaller WEF-matrix than the
ResNet-based models or LeNet, which may reduce
the amount of information available for distinguishing benign and
free-rider behavior.
In combination with non-IID data heterogeneity, this can make the
separation between benign and suspicious WEF patterns more difficult.

Finally, the performance of STD-DAGMM in our experiments differs from
that reported in \cite{STD-DAGMM}.
One possible reason is the evaluation protocol used in this study.
STD-DAGMM requires pre-training using benign model updates, whereas in
our evaluation the client data are redistributed using a different
random seed between detector pre-training and the subsequent detection
experiment.
This introduces a distribution shift between the behavior observed
during pre-training and that observed during evaluation, which may
partly explain the performance degradation.
This result also highlights a practical limitation of pre-training-based
detectors when the client data distribution changes over time.
\begin{table}[t]
\caption{F1-score comparison on MNIST and Adult.}
\label{tab:f1_main_mnist_adult}
\centering
\scriptsize
\setlength{\tabcolsep}{2.0pt}
\renewcommand{\arraystretch}{0.90}
\begin{tabular}{@{}lllc c c@{}}
\toprule
Dataset & Dist. & Attack & STD-DAGMM & WEF-na & S2-WEF \\
\midrule
\multirow{10}{*}{\rotatebox[origin=c]{90}{MNIST}} & \multirow{5}{*}{IID} & RWA & $0.588\!\pm\!0.537$ & \textbf{$0.980\!\pm\!0.000$} & $0.978\!\pm\!0.011$ \\
 &  & SPA & $0.371\!\pm\!0.432$ & \textbf{$0.980\!\pm\!0.000$} & $0.978\!\pm\!0.011$ \\
 &  & DWA & $0.222\!\pm\!0.185$ & $0.298\!\pm\!0.420$ & \textbf{$0.978\!\pm\!0.011$} \\
 &  & ADWA & $0.338\!\pm\!0.392$ & \textbf{$0.978\!\pm\!0.004$} & \textbf{$0.978\!\pm\!0.011$} \\
 &  & AWCA & $0.352\!\pm\!0.302$ & $0.196\!\pm\!0.438$ & \textbf{$0.978\!\pm\!0.011$} \\
\cmidrule(lr){2-6}
 & \multirow{5}{*}{Non-IID} & RWA & $0.976\!\pm\!0.015$ & $0.976\!\pm\!0.009$ & \textbf{$0.992\!\pm\!0.009$} \\
 &  & SPA & $0.185\!\pm\!0.169$ & $0.976\!\pm\!0.009$ & \textbf{$0.992\!\pm\!0.009$} \\
 &  & DWA & $0.108\!\pm\!0.103$ & $0.184\!\pm\!0.174$ & \textbf{$0.992\!\pm\!0.009$} \\
 &  & ADWA & $0.299\!\pm\!0.234$ & $0.891\!\pm\!0.016$ & \textbf{$0.992\!\pm\!0.009$} \\
 &  & AWCA & $0.259\!\pm\!0.278$ & $0.008\!\pm\!0.018$ & \textbf{$0.992\!\pm\!0.009$} \\
\midrule
\multirow{10}{*}{\rotatebox[origin=c]{90}{Adult}} & \multirow{5}{*}{IID} & RWA & $0.373\!\pm\!0.415$ & $0.972\!\pm\!0.013$ & \textbf{$0.992\!\pm\!0.009$} \\
 &  & SPA & $0.970\!\pm\!0.033$ & $0.972\!\pm\!0.013$ & \textbf{$0.992\!\pm\!0.009$} \\
 &  & DWA & $0.075\!\pm\!0.068$ & $0.390\!\pm\!0.534$ & \textbf{$0.992\!\pm\!0.009$} \\
 &  & ADWA & $0.470\!\pm\!0.289$ & $0.972\!\pm\!0.013$ & \textbf{$0.992\!\pm\!0.009$} \\
 &  & AWCA & $0.127\!\pm\!0.149$ & $0.390\!\pm\!0.534$ & \textbf{$0.992\!\pm\!0.009$} \\
\cmidrule(lr){2-6}
 & \multirow{5}{*}{Non-IID} & RWA & $0.521\!\pm\!0.406$ & \textbf{$0.980\!\pm\!0.000$} & $0.927\!\pm\!0.119$ \\
 &  & SPA & $0.534\!\pm\!0.318$ & \textbf{$0.980\!\pm\!0.000$} & $0.861\!\pm\!0.265$ \\
 &  & DWA & $0.024\!\pm\!0.028$ & $0.000\!\pm\!0.000$ & \textbf{$0.980\!\pm\!0.007$} \\
 &  & ADWA & $0.578\!\pm\!0.302$ & \textbf{$0.980\!\pm\!0.000$} & $0.918\!\pm\!0.137$ \\
 &  & AWCA & $0.144\!\pm\!0.142$ & $0.000\!\pm\!0.000$ & \textbf{$0.919\!\pm\!0.100$} \\
\bottomrule
\end{tabular}
\vspace{1pt}
\begin{minipage}{\columnwidth}
\scriptsize
Values are mean $\pm$ sample standard deviation over five random seeds.
\textit{Abbreviations:} WEF-na, WEF-Defense with no accumulation of the WEF matrix.
\end{minipage}
\end{table}

\begin{table}[t]
\caption{F1-score comparison on CIFAR-10 and CIFAR-100.}
\label{tab:f1_main_cifar}
\centering
\scriptsize
\setlength{\tabcolsep}{2.0pt}
\renewcommand{\arraystretch}{0.90}
\begin{tabular}{@{}lllc c c@{}}
\toprule
Dataset & Dist. & Attack & STD-DAGMM & WEF-na & S2-WEF \\
\midrule
\multirow{10}{*}{\rotatebox[origin=c]{90}{CIFAR-10}} & \multirow{5}{*}{IID} & RWA & $0.788\!\pm\!0.414$ & $0.985\!\pm\!0.003$ & \textbf{$0.986\!\pm\!0.007$} \\
 &  & SPA & $0.766\!\pm\!0.428$ & $0.985\!\pm\!0.003$ & \textbf{$0.986\!\pm\!0.007$} \\
 &  & DWA & $0.738\!\pm\!0.415$ & $0.985\!\pm\!0.003$ & \textbf{$0.986\!\pm\!0.007$} \\
 &  & ADWA & $0.467\!\pm\!0.381$ & $0.985\!\pm\!0.003$ & \textbf{$0.986\!\pm\!0.007$} \\
 &  & AWCA & $0.738\!\pm\!0.415$ & $0.985\!\pm\!0.003$ & \textbf{$0.986\!\pm\!0.007$} \\
\cmidrule(lr){2-6}
 & \multirow{5}{*}{Non-IID} & RWA & $0.957\!\pm\!0.052$ & \textbf{$0.984\!\pm\!0.006$} & \textbf{$0.984\!\pm\!0.015$} \\
 &  & SPA & $0.630\!\pm\!0.395$ & \textbf{$0.984\!\pm\!0.006$} & \textbf{$0.984\!\pm\!0.015$} \\
 &  & DWA & $0.583\!\pm\!0.526$ & $0.000\!\pm\!0.000$ & \textbf{$0.984\!\pm\!0.015$} \\
 &  & ADWA & $0.386\!\pm\!0.361$ & $0.969\!\pm\!0.028$ & \textbf{$0.980\!\pm\!0.019$} \\
 &  & AWCA & $0.566\!\pm\!0.518$ & $0.000\!\pm\!0.000$ & \textbf{$0.984\!\pm\!0.015$} \\
\midrule
\multirow{10}{*}{\rotatebox[origin=c]{90}{CIFAR-100}} & \multirow{5}{*}{IID} & RWA & $0.971\!\pm\!0.024$ & \textbf{$0.990\!\pm\!0.000$} & $0.988\!\pm\!0.013$ \\
 &  & SPA & $0.664\!\pm\!0.352$ & \textbf{$0.990\!\pm\!0.000$} & $0.988\!\pm\!0.013$ \\
 &  & DWA & $0.843\!\pm\!0.036$ & \textbf{$0.990\!\pm\!0.000$} & $0.988\!\pm\!0.013$ \\
 &  & ADWA & $0.872\!\pm\!0.073$ & \textbf{$0.990\!\pm\!0.000$} & $0.988\!\pm\!0.013$ \\
 &  & AWCA & $0.837\!\pm\!0.027$ & \textbf{$0.990\!\pm\!0.000$} & $0.988\!\pm\!0.013$ \\
\cmidrule(lr){2-6}
 & \multirow{5}{*}{Non-IID} & RWA & $0.982\!\pm\!0.012$ & \textbf{$0.989\!\pm\!0.002$} & \textbf{$0.989\!\pm\!0.002$} \\
 &  & SPA & $0.760\!\pm\!0.425$ & \textbf{$0.989\!\pm\!0.002$} & \textbf{$0.989\!\pm\!0.002$} \\
 &  & DWA & $0.749\!\pm\!0.421$ & \textbf{$0.989\!\pm\!0.002$} & \textbf{$0.989\!\pm\!0.002$} \\
 &  & ADWA & $0.750\!\pm\!0.420$ & \textbf{$0.989\!\pm\!0.002$} & \textbf{$0.989\!\pm\!0.002$} \\
 &  & AWCA & $0.752\!\pm\!0.422$ & \textbf{$0.989\!\pm\!0.002$} & \textbf{$0.989\!\pm\!0.002$} \\
\bottomrule
\end{tabular}
\vspace{1pt}
\begin{minipage}{\columnwidth}
\scriptsize
Values are mean $\pm$ sample standard deviation over five random seeds.
\textit{Abbreviations:} WEF-na, WEF-Defense with no accumulation of the WEF matrix.
\end{minipage}
\end{table}

\subsection{Ablation Study}
\label{sec:ablation}
%long version
We conduct an ablation study on Adult and CIFAR-10 to evaluate the
contributions of the deviation score $\mathrm{Dev}$, the similarity score
$\gamma$, the $L^2$-distance in $\gamma$, and the vote-based verification.
We use AWCA as a representative global-model-mimicking attack and SPA as a
perturbation-based attack. 
%Unless otherwise specified, the settings follow
%Sec.~\ref{sec:main-experiment}, with five random seeds.
We compare $\mathrm{Dev}$ only, $\gamma$ only with cosine similarity,
$\gamma$ only with cosine similarity divided by the $L^2$-distance,
clustering only, and the full S2-WEF pipeline.
The results are shown in Tab.~\ref{tab:ablation_precision_recall_fpr}.

The results demonstrate that $\mathrm{Dev}$ and $\gamma$ capture
complementary attack characteristics.
For SPA, $\mathrm{Dev}$ alone achieves a recall of $1.000$ on both datasets,
whereas both $\gamma$-only variants obtain zero recall.
In contrast, for AWCA on Adult, $\mathrm{Dev}$ alone fails completely,
while $\gamma$ only (cos/$\ell_2$) achieves a recall of
$0.996\pm0.009$.
This confirms that $\mathrm{Dev}$ is effective for perturbation-based
attacks, whereas $\gamma$ is effective for global-model-mimicking attacks.

The $L^2$-distance also improves the robustness of $\gamma$.
On Adult/AWCA, recall increases from $0.242\pm0.429$ with cosine similarity
alone to $0.996\pm0.009$ with cos/$\ell_2$.
On CIFAR-10, where cosine similarity already performs well, the improvement
is smaller but precision increases from $0.964\pm0.027$ to
$0.975\pm0.015$.

Finally, the full pipeline reduces false positives compared with clustering
alone while maintaining the same recall in the evaluated attack settings.
This effect is particularly clear when no free-riders are present:
as shown in Tab.~\ref{tab:no_attack_fpr}, FPR decreases from
$0.176\pm0.128$ to $0.088\pm0.050$ on Adult and from
$0.178\pm0.036$ to $0.097\pm0.016$ on CIFAR-10.
These results support the use of threshold-based vote verification to
suppress spurious suspicious clusters.
\iffalse
%short version
We conduct an ablation study on Adult and CIFAR-10 (non-IID) using AWCA and SPA
to evaluate the contributions of $\mathrm{Dev}$, $\gamma$, the
$L^2$-distance in $\gamma$, and vote-based verification.
We compare $\mathrm{Dev}$ only, $\gamma$ only with cosine similarity,
$\gamma$ only with cos/$\ell_2$, clustering only, and the full S2-WEF
pipeline under the settings of Sec.~\ref{sec:main-experiment} with five
random seeds.
The results are shown in Tab.~\ref{tab:ablation_precision_recall_fpr}.

The results confirm that $\mathrm{Dev}$ and $\gamma$ capture complementary
attack characteristics.
For SPA, $\mathrm{Dev}$ alone achieves a recall of $1.000$ on both
datasets, whereas the $\gamma$-only variants obtain zero recall.
Conversely, on Adult/AWCA, $\mathrm{Dev}$ fails, while $\gamma$ with
cos/$\ell_2$ achieves a recall of $0.996\pm0.009$.
The $L^2$-distance is also important: on Adult/AWCA, it increases the
recall of $\gamma$ from $0.242\pm0.429$ to $0.996\pm0.009$.

Finally, vote-based verification reduces false positives without reducing
recall in the evaluated attack settings.
Without free-riders, it reduces FPR from $0.176\pm0.128$ to
$0.088\pm0.050$ on Adult and from $0.178\pm0.036$ to
$0.097\pm0.016$ on CIFAR-10 (Tab.~\ref{tab:no_attack_fpr}),
supporting its role in suppressing spurious suspicious clusters.

\fi
\begin{table}[t]
\caption{Ablation results of S2-WEF.}
\label{tab:ablation_precision_recall_fpr}
\centering
\scriptsize
\setlength{\tabcolsep}{1.7pt}
\renewcommand{\arraystretch}{0.90}
\begin{tabular}{@{}lllccc@{}}
\toprule
Dataset & Attack & Variant & Precision & Recall & FPR \\
\midrule
\multirow{10}{*}{Adult} & \multirow{5}{*}{AWCA} & $Dev$ only & $0.000\!\pm\!0.000$ & $0.000\!\pm\!0.000$ & $0.154\!\pm\!0.063$ \\
 &  & $\gamma$ only (cos) & $0.660\!\pm\!0.422$ & $0.242\!\pm\!0.429$ & $0.022\!\pm\!0.047$ \\
 &  & $\gamma$ only (cos/$\ell_2$) & $0.757\!\pm\!0.210$ & $0.996\!\pm\!0.009$ & $0.043\!\pm\!0.044$ \\
 &  & Clustering only & $0.873\!\pm\!0.099$ & $0.954\!\pm\!0.102$ & $0.015\!\pm\!0.012$ \\
 &  & Full pipeline & $0.888\!\pm\!0.110$ & $0.954\!\pm\!0.102$ & $0.013\!\pm\!0.013$ \\
\cmidrule(lr){2-6}
 & \multirow{5}{*}{SPA} & $Dev$ only & $0.960\!\pm\!0.000$ & $1.000\!\pm\!0.000$ & $0.004\!\pm\!0.000$ \\
 &  & $\gamma$ only (cos) & $0.000\!\pm\!0.000$ & $0.000\!\pm\!0.000$ & $0.022\!\pm\!0.047$ \\
 &  & $\gamma$ only (cos/$\ell_2$) & $0.000\!\pm\!0.000$ & $0.000\!\pm\!0.000$ & $0.070\!\pm\!0.055$ \\
 &  & Clustering only & $0.829\!\pm\!0.252$ & $0.879\!\pm\!0.270$ & $0.019\!\pm\!0.028$ \\
 &  & Full pipeline & $0.844\!\pm\!0.260$ & $0.879\!\pm\!0.270$ & $0.017\!\pm\!0.029$ \\
\midrule
\multirow{10}{*}{CIFAR-10} & \multirow{5}{*}{AWCA} & $Dev$ only & $0.824\!\pm\!0.304$ & $0.856\!\pm\!0.307$ & $0.021\!\pm\!0.036$ \\
 &  & $\gamma$ only (cos) & $0.964\!\pm\!0.027$ & $1.000\!\pm\!0.000$ & $0.004\!\pm\!0.003$ \\
 &  & $\gamma$ only (cos/$\ell_2$) & $0.975\!\pm\!0.015$ & $1.000\!\pm\!0.000$ & $0.003\!\pm\!0.002$ \\
 &  & Clustering only & $0.944\!\pm\!0.010$ & $1.000\!\pm\!0.000$ & $0.006\!\pm\!0.001$ \\
 &  & Full pipeline & $0.968\!\pm\!0.029$ & $1.000\!\pm\!0.000$ & $0.004\!\pm\!0.003$ \\
\cmidrule(lr){2-6}
 & \multirow{5}{*}{SPA} & $Dev$ only & $0.970\!\pm\!0.007$ & $1.000\!\pm\!0.000$ & $0.003\!\pm\!0.001$ \\
 &  & $\gamma$ only (cos) & $0.000\!\pm\!0.000$ & $0.000\!\pm\!0.000$ & $0.005\!\pm\!0.006$ \\
 &  & $\gamma$ only (cos/$\ell_2$) & $0.000\!\pm\!0.000$ & $0.000\!\pm\!0.000$ & $0.007\!\pm\!0.003$ \\
 &  & Clustering only & $0.944\!\pm\!0.010$ & $1.000\!\pm\!0.000$ & $0.006\!\pm\!0.001$ \\
 &  & Full pipeline & $0.968\!\pm\!0.029$ & $1.000\!\pm\!0.000$ & $0.004\!\pm\!0.003$ \\
\bottomrule
\end{tabular}
\vspace{1pt}
\begin{minipage}{\columnwidth}
\scriptsize
Values are mean $\pm$ sample standard deviation over five random seeds.
\end{minipage}
\end{table}

\begin{table}[t]
    \caption{False-positive rate without free riders.}
    \label{tab:no_attack_fpr}
    \centering
    \scriptsize 
    \setlength{\tabcolsep}{4.0pt} 
    \renewcommand{\arraystretch}{0.95} 
    \begin{tabular}{@{}lcc@{}} 
        \toprule Dataset & Clustering only & Full pipeline \\ \midrule Adult & $0.176\!\pm\!0.128$ & $0.088\!\pm\!0.050$ \\ CIFAR-10 & $0.178\!\pm\!0.036$ & $0.097\!\pm\!0.016$ \\ \bottomrule
    \end{tabular}
    \vspace{1pt}
    \begin{minipage}{\columnwidth}
    \scriptsize
    Values are mean $\pm$ sample standard deviation over five random seeds.
    \end{minipage}
\end{table}

\subsection{Sensitivity Analysis of the Similarity-Score Threshold}
\label{sec:gamma-sensitivity}
We analyze the sensitivity to $k$ in (\ref{eq:gamma-thr-value}) on
Adult and CIFAR-10 (non-IID) with free-rider ratios of $10\%$ and $30\%$, varying
$k\in\{1,2,3,4,5,7,10\}$ for both $\gamma$-only detection and the full
S2-WEF pipeline (Tab.~\ref{tab:gamma_sensitivity_single_column}).

For $\gamma$-only detection, small $k$ values tend to preserve recall but
reduce precision because the threshold becomes less restrictive.
On CIFAR-10, performance largely saturates for $k\geq5$.
On Adult, increasing $k$ also improves precision, but excessively large
values reduce recall under the $30\%$ free-rider setting; for example,
the F1-score decreases from $0.963$ at $k=5$ to $0.704$ at $k=10$.
We therefore use $k=5$ in the main experiments as a conservative choice
that improves precision while avoiding unnecessary loss of recall.

The full S2-WEF pipeline is substantially less sensitive to $k$ over the
moderate range $k=1$--$5$.
For example, its F1-score remains unchanged at $0.953$ on Adult with
$10\%$ free-riders despite the large variation observed for $\gamma$ alone.
This indicates that combining score-wise thresholding with clustering and
vote-based verification reduces the dependence of the final decision on the
specific choice of $k$.

\begin{table}[t]
\caption{Sensitivity of the $\gamma$-threshold parameter $k$.}
\label{tab:gamma_sensitivity_single_column}
\centering
\scriptsize
\setlength{\tabcolsep}{1.65pt}
\renewcommand{\arraystretch}{0.88}
\begin{tabular}{@{}llc@{\hspace{3pt}}ccc@{\hspace{5pt}}ccc@{}}
\toprule
& & & \multicolumn{3}{c}{$\gamma$ only} & \multicolumn{3}{c}{Full pipeline} \\
\cmidrule(lr){4-6}\cmidrule(lr){7-9}
Dataset & FR & $k$ & Precision & Recall & F1 & Precision & Recall & F1 \\
\midrule
\multirow{14}{*}{Adult} & \multirow{7}{*}{10\%} & 1 & 0.399 & 1.000 & 0.567 & 0.913 & 1.000 & 0.953 \\
 &  & 2 & 0.481 & 1.000 & 0.649 & 0.913 & 1.000 & 0.953 \\
 &  & 3 & 0.556 & 1.000 & 0.713 & 0.913 & 1.000 & 0.953 \\
 &  & 4 & 0.694 & 0.993 & 0.805 & 0.913 & 1.000 & 0.953 \\
 &  & 5 & 0.777 & 0.993 & 0.865 & 0.913 & 1.000 & 0.953 \\
 &  & 7 & 0.931 & 1.000 & 0.964 & 0.913 & 1.000 & 0.953 \\
 &  & 10 & 0.974 & 1.000 & 0.986 & 0.919 & 0.993 & 0.953 \\
\cmidrule(lr){2-9}
 & \multirow{7}{*}{30\%} & 1 & 0.774 & 1.000 & 0.871 & 0.971 & 1.000 & 0.985 \\
 &  & 2 & 0.832 & 1.000 & 0.904 & 0.971 & 1.000 & 0.985 \\
 &  & 3 & 0.872 & 1.000 & 0.929 & 0.971 & 1.000 & 0.985 \\
 &  & 4 & 0.906 & 0.993 & 0.944 & 0.971 & 1.000 & 0.985 \\
 &  & 5 & 0.938 & 0.993 & 0.963 & 0.971 & 1.000 & 0.985 \\
 &  & 7 & 0.973 & 0.919 & 0.941 & 0.973 & 0.917 & 0.941 \\
 &  & 10 & 0.964 & 0.688 & 0.704 & 0.829 & 0.674 & 0.678 \\
\midrule
\multirow{14}{*}{CIFAR-10} & \multirow{7}{*}{10\%} & 1 & 0.393 & 1.000 & 0.564 & 0.964 & 1.000 & 0.981 \\
 &  & 2 & 0.759 & 1.000 & 0.862 & 0.964 & 1.000 & 0.981 \\
 &  & 3 & 0.864 & 1.000 & 0.927 & 0.964 & 1.000 & 0.981 \\
 &  & 4 & 0.947 & 1.000 & 0.973 & 0.964 & 1.000 & 0.981 \\
 &  & 5 & 0.971 & 1.000 & 0.985 & 0.964 & 1.000 & 0.981 \\
 &  & 7 & 1.000 & 1.000 & 1.000 & 0.964 & 1.000 & 0.981 \\
 &  & 10 & 1.000 & 1.000 & 1.000 & 0.964 & 1.000 & 0.981 \\
\cmidrule(lr){2-9}
 & \multirow{7}{*}{30\%} & 1 & 0.934 & 1.000 & 0.966 & 0.980 & 1.000 & 0.990 \\
 &  & 2 & 0.993 & 1.000 & 0.996 & 0.985 & 1.000 & 0.992 \\
 &  & 3 & 1.000 & 1.000 & 1.000 & 0.985 & 1.000 & 0.992 \\
 &  & 4 & 1.000 & 1.000 & 1.000 & 0.985 & 1.000 & 0.992 \\
 &  & 5 & 1.000 & 1.000 & 1.000 & 0.985 & 1.000 & 0.992 \\
 &  & 7 & 1.000 & 1.000 & 1.000 & 0.985 & 1.000 & 0.992 \\
 &  & 10 & 1.000 & 1.000 & 1.000 & 0.985 & 1.000 & 0.992 \\
\bottomrule
\end{tabular}
\vspace{1pt}
\begin{minipage}{\columnwidth}
\scriptsize
All values are averages over three random seeds.
\end{minipage}
\end{table}

\subsection{Robustness to Attack Scenarios and Free-Rider Ratios}
\label{sec:attack-robustness}
We evaluate S2-WEF under different attack scenarios and free-rider ratios
using AWCA on Adult and CIFAR-10 (non-IID).
In addition to the main \emph{Dynamic} setting (Sec.~\ref{sec:main-experiment}), we consider
\emph{Dynamic-random}, where the attacking client is randomly selected at
each round from the third global communication round onward.
Over three random seeds, the F1-scores for Dynamic and Dynamic-random are
$0.953$ and $0.958$ on Adult, and $0.981$ and $0.990$ on CIFAR-10,
respectively, showing that S2-WEF remains effective even when the attacking client changes across rounds.
%respectively, showing that S2-WEF remains robust regardless of the onset timing and frequency of free-riding.

We also vary the free-rider ratio from $10\%$ to $40\%$ within the
honest-majority setting.
As shown in Tab.~\ref{tab:free_rider_ratio_change}, S2-WEF achieves
F1-scores above $0.95$ in all settings.
Recall is $1.000$ except on Adult at $40\%$, where it is $0.972$,
indicating robustness to different free-rider ratios.

\begin{table}[t] 
    \caption{Detection performance under different free-rider ratios.} \label{tab:free_rider_ratio_change} 
    \centering 
    \scriptsize 
    \setlength{\tabcolsep}{4.0pt} 
    \renewcommand{\arraystretch}{0.95} 
    \begin{tabular}{@{}lcccc@{}} 
        \toprule Dataset & FR ratio & Precision & Recall & F1 \\ 
        \midrule 
        \multirow{4}{*}{Adult} & 10\% & 0.913 & 1.000 & 0.953 \\ & 20\% & 0.976 & 1.000 & 0.988 \\ & 30\% & 0.971 & 1.000 & 0.985 \\ & 40\% & 0.982 & 0.972 & 0.977 \\ 
        \midrule 
        \multirow{4}{*}{CIFAR-10} & 10\% & 0.964 & 1.000 & 0.981 \\ & 20\% & 0.979 & 1.000 & 0.989 \\ & 30\% & 0.985 & 1.000 & 0.992 \\ & 40\% & 0.996 & 1.000 & 0.998 \\ 
        \bottomrule 
    \end{tabular}
    \vspace{1pt}
\begin{minipage}{\columnwidth}
\scriptsize
All values are averages over three random seeds.
\end{minipage}
\end{table}

\subsection{Scalability with the Number of Clients}
\label{sec:client-scalability}
We evaluate the detection performance of S2-WEF as the number of clients
increases using CIFAR-10 and CIFAR-100 (non-IID) under AWCA.
We vary the number of clients as $N\in\{5,10,20,50\}$.
The free-rider ratio is set to $10\%$ for $N=10,20,$ and $50$.
For $N=5$, one free-rider is used, corresponding to a ratio of $20\%$,
since at least one malicious client is required.
All results are averaged over three random seeds.

As shown in Tab.~\ref{tab:client_count}, S2-WEF maintains high detection
performance across all client counts.
Recall remains $1.000$ in every setting, and the F1-score is at least
$0.979$ on CIFAR-10 and $0.985$ on CIFAR-100.
Moreover, increasing the number of clients does not degrade detection
performance; S2-WEF achieves F1-scores of $0.997$ and $1.000$ with
$50$ clients on CIFAR-10 and CIFAR-100, respectively.
These results indicate that the detection effectiveness of S2-WEF remains
stable as the number of participating clients increases from $5$ to $50$.

\begin{table}[t]
\caption{Detection performance with different numbers of clients.}
\label{tab:client_count}
\centering
\scriptsize
\setlength{\tabcolsep}{2.5pt}
\renewcommand{\arraystretch}{0.95}
\begin{tabular}{@{}lcccccc@{}}
\toprule
Dataset & Clients & FRs & FR ratio & Precision & Recall & F1 \\
\midrule
\multirow{4}{*}{CIFAR-10} & 5 & 1 & 20\% & 0.959 & 1.000 & 0.979 \\
 & 10 & 1 & 10\% & 0.964 & 1.000 & 0.981 \\
 & 20 & 2 & 10\% & 0.983 & 1.000 & 0.992 \\
 & 50 & 5 & 10\% & 0.994 & 1.000 & 0.997 \\
\midrule
\multirow{4}{*}{CIFAR-100} & 5 & 1 & 20\% & 0.987 & 1.000 & 0.994 \\
 & 10 & 1 & 10\% & 0.971 & 1.000 & 0.985 \\
 & 20 & 2 & 10\% & 0.975 & 1.000 & 0.987 \\
 & 50 & 5 & 10\% & 0.999 & 1.000 & 1.000 \\
\bottomrule
\end{tabular}
\vspace{1pt}
\begin{minipage}{\columnwidth}
\scriptsize
FRs denotes the number of free riders. The five-client setting uses one free rider (20\%), while the other settings use a 10\% free-rider ratio. All values are averages over three random seeds.
\end{minipage}
\end{table}

\subsection{Impact on Global Model Accuracy}
\label{sec:main-task-accuracy}
\iffalse
Finally, we evaluate whether S2-WEF affects the main-task accuracy of the
global model.
We compare three settings on Adult and CIFAR-10: FL without free-riders or
defense, FL without free-riders using S2-WEF, and FL with $10\%$ free-riders
performing AWCA using S2-WEF.
All results in Tab.~\ref{tab:global_model_main_task_accuracy} are averages
over three random seeds.

The global-model accuracy remains similar across the three settings.
Without free-riders, applying S2-WEF changes the accuracy from $63.962\%$ to
$64.260\%$ on Adult and from $90.163\%$ to $90.160\%$ on CIFAR-10,
indicating little impact from the defense itself.
Even with $10\%$ AWCA free-riders, the accuracy remains $64.316\%$ on Adult
and $89.753\%$ on CIFAR-10.
Thus, the observed differences from the no-attack/no-defense setting are at
most $0.41$ percentage points, showing that S2-WEF maintains the main-task
performance of the global model under the evaluated settings.
\fi
We evaluate the impact of S2-WEF on global-model accuracy on Adult and
CIFAR-10 under non-IID data using three settings: no attack/no defense,
no attack with S2-WEF, and $10\%$ AWCA free-riders with S2-WEF.
As shown in Tab.~\ref{tab:global_model_main_task_accuracy}, mean accuracy
remains similar across all settings, including those with AWCA free-riders.
The differences from the no-attack/no-defense setting are at most $1.09$
percentage points, suggesting limited impact on main-task performance.

\begin{table}[t]
    \caption{Global-model main-task accuracy under different attack and defense settings.}
    \label{tab:global_model_main_task_accuracy}
    \centering 
    \scriptsize 
    \setlength{\tabcolsep}{4.0pt} 
    \renewcommand{\arraystretch}{0.95} 
    \begin{tabular}{@{}llc@{}} 
        \toprule 
        Dataset & Setting & Accuracy (\%) \\ 
        \midrule 
        \multirow{3}{*}{Adult} & No attack, no defense & $62.637\!\pm\!7.027$ \\
         & No attack, S2-WEF & $62.269\!\pm\!7.688$ \\ 
         & 10\% attackers, S2-WEF & $63.730\!\pm\!7.265$ \\ 
        \midrule 
        \multirow{3}{*}{CIFAR-10} & No attack, no defense & $90.294\!\pm\!0.603$ \\ 
        & No attack, S2-WEF & $90.068\!\pm\!0.550$ \\ 
        & 10\% attackers, S2-WEF & $89.734\!\pm\!0.401$ \\ 
        \bottomrule 
    \end{tabular} 
    \vspace{1pt} 
    \begin{minipage}
        {\columnwidth} \scriptsize Values are mean $\pm$ sample standard deviation over five random seeds. 
    \end{minipage} 
\end{table}

\section{Limitations and future work}\label{sec:LimandDis}
\paragraph{Heterogeneous FL.}
In practical deployments, clients can be heterogeneous in both data distributions and model architectures.
Under strong data heterogeneity, performance can drop (e.g., Adult under non-IID in Tab.~\ref{tab:f1_main_mnist_adult}), and combining S2-WEF with non-IID-aware FL methods such as FedProx~\cite{Fedprox} or FedPer~\cite{Fedper} is a promising direction.
Architecture heterogeneity is more fundamental: S2-WEF assumes a consistent penultimate-layer shape to construct comparable WEF-matrices, and extending it to heterogeneous architectures (e.g., via a shared representation) remains open.

\paragraph{Fairness and incentive mechanisms}
\iffalse
In S2-WEF, the models submitted by clients identified as free-riders are excluded to maintain the quality of the global model during the model aggregation process. 
However, unlike some previous studies, all clients download the same global model regardless of the detection results from the previous global communication round. 
While this strategy enables accurate detection of free-riders in each round, another perspective may claim that this method could not maintain fairness in FL. 
From this perspective, we should prepare methods to ensure fairness, and incentive mechanisms are likely suitable tools for this purpose.
Since many incentive mechanisms for FL have already been proposed \cite{IncentiveMechanism-survey}, we believe a new approach is required to combine these mechanisms with S2-WEF.
\fi
In S2-WEF, updates from clients detected as free-riders are excluded from aggregation to preserve global model quality. 
However, unlike some prior work, all clients download the same global model regardless of the previous round's detection outcome. 
While this choice supports accurate round-wise detection, it may raise fairness concerns in FL. 
Therefore, designing incentive mechanisms that align with S2-WEF is an important direction. 
Although many FL incentive mechanisms have been studied~\cite{IncentiveMechanism-survey}, a tailored approach is needed to integrate them with our detection pipeline.

\paragraph{Learning the WEF-matrix itself.}
A remaining challenge is an adaptive strategy in which a client first behaves benignly and later synthesizes counterfeit WEF-matrices by learning previously observed benign-like WEF patterns.
One mitigation is to randomize the layer (or subset of layers) used to compute the WEF-matrix, forcing attackers to anticipate multiple possible WEF representations.
In principle, an attacker could pre-train on local data to generate valid WEF-matrices for many layers; however, for deep and large models, doing so broadly can be computationally expensive, making honest participation a more rational choice.
\iffalse
Another challenge arises when benign clients turn into free-riders as they could generate new counterfeit WEF-matrices using previous valid WEF-matrices. 
While we do not fully address this attack, one mitigation strategy involves randomly changing the layer responsible for calculating the WEF matrix.
One approach free-riders could take to adapt to the central server's random layer changes is to pre-train using their local data to generate many valid WEF-matrices, enabling them to counterfeit WEF-matrices for all layers.
However, especially for massive networks with very deep layers, considering the computational resources required to do this, it would be a rational decision to participate in FL honestly.
\fi
\paragraph{Federated LLMs}
Applying our method to federated large language models (Federated LLMs) is of great interest as a future direction. 
Recently, approaches using FL for fine-tuning and prompt learning of LLMs have gained attention, and many methods have been proposed \cite{FederatedLLM-survey}. 
We believe that addressing free-riders is also essential in this field.

\section{Conclusion}\label{sec:Conclusion}
In this paper, we proposed S2-WEF for detecting dynamic free-riders in
cross-silo FL.
S2-WEF combines server-side simulation of global-model-mimicking attack
patterns with client-to-client deviation to enable round-wise detection.
Experiments on four datasets and five attacks show that S2-WEF enables
accurate detection without proxy datasets or pre-training.
We leave incentive integration and extensions to heterogeneous FL as future work.
%\section{LLM usage considerations}
%LLMs were used for editorial purposes in this manuscript, and all outputs were inspected by the authors to ensure accuracy and originality.

\section*{Acknowledgments}
The author is deeply grateful to Takeru Fukuoka for leading the project and for providing continuous support from the early stage to the finalization of this work, including insightful advice on the proposed method and experiments, fruitful discussions, and detailed reviews of the manuscript. 
The author also thanks Haber Janosch for early discussions on risks in FL that inspired this research direction, and for his comments on the abstract and the introduction. 
The author thanks Yoshiki Higashikado for valuable discussions and for setting up the experimental environment. 
The author also thanks Takuma Takeuchi, Akira Ito, and Takahide Matsutsuka for their discussions and guidance on the research direction.

\bibliographystyle{IEEEtran}
\bibliography{IEEEabrv,reference}

\end{document}